\documentclass{article}
\usepackage[final]{neurips_2022}

\usepackage{natbib}
\setcitestyle{numbers}
\usepackage[utf8]{inputenc} 
\usepackage[T1]{fontenc}    
\usepackage{hyperref}       
\usepackage{url}            
\usepackage{booktabs}       
\usepackage{amsfonts}       
\usepackage{nicefrac}       
\usepackage{microtype}      
\usepackage{xcolor,colortbl}         

\usepackage{tabularx}
\usepackage{adjustbox}
\usepackage{xspace}
\usepackage{enumitem}
\usepackage{amssymb}
\usepackage{amsmath}
\usepackage{multirow}
\usepackage{graphicx}
\usepackage[center]{subfigure}
\usepackage{wrapfig}
\usepackage[]{algorithmic} 
\usepackage{tablefootnote}
\usepackage{threeparttable}
\usepackage{makecell}
\usepackage[font={small}]{caption}
\usepackage{diagbox}
\usepackage{empheq}
\usepackage{amsthm}
\usepackage{todonotes}
\usepackage{algorithm}

\title{Zero-shot Transfer Learning within a Heterogeneous Graph via Knowledge Transfer Networks}

\author{%
  Minji Yoon\thanks{Work done while interning at Google} \\
  Carnegie Mellon University\\
  \And
  John Palowitch\\
  Google Research\\
  \And
  Dustin Zelle\\
  Google Research\\
  \And
  Ziniu Hu$^*$\\
  University of California Los Angeles
  \And
  Ruslan Salakhutdinov\\
  Carnegie Mellon University\\
  \And
  Bryan Perozzi\\
  Google Research\\
}

\begin{document}
\maketitle

\newcommand{\method}{\textsc{KTN}\xspace}
\renewcommand{\qedsymbol}{$\blacksquare$}

\newtheorem{problem_inf}{Informal Problem Definition}
\newtheorem{problem}{Problem Definition}
\newtheorem{theorem}{Theorem}
\newtheorem{obs}{Observation}

\newcommand{\john}[1]{\todo[inline,color=green!20!white]{\textbf{John:} #1}}
\newcommand{\bryan}[1]{\todo[inline,color=blue!20!white]{\textbf{Bryan:} #1}}
\newcommand{\dustin}[1]{\todo[inline,color=red!20!white]{\textbf{Dustin:} #1}}
\newcommand{\minji}[1]{\textcolor{blue}{#1}}

\newcommand{\concat}{\mathbin\Vert}

\newcommand{\snre}{\sigma_e}
\newcommand{\snrf}{\sigma_f}

\begin{abstract}
Data continuously emitted from industrial ecosystems such as social or e-commerce platforms are commonly represented as heterogeneous graphs (HG) composed of multiple node/edge types. 
State-of-the-art graph learning methods for HGs known as heterogeneous graph neural networks (HGNNs) are applied to learn deep context-informed node representations.
However, many HG datasets from industrial applications suffer from label imbalance between node types.
As there is no direct way to learn using labels rooted at different node types, HGNNs have been applied on only a few node types with abundant labels.
We propose a zero-shot transfer learning module for HGNNs called a Knowledge Transfer Network (\method) that transfers knowledge from \emph{label-abundant} node types to \emph{zero-labeled} node types through rich relational information given in the HG.
\method is derived from the theoretical relationship, which we introduce in this work, between distinct \emph{feature extractors} for each node types given in a HGNN model.
\method improves performance of $6$ different types of HGNN models by up to $960\%$ for inference on zero-labeled node types and outperforms state-of-the-art transfer learning baselines by up to $73\%$ across $18$ different transfer learning tasks on HGs.
\end{abstract}

\section{Introduction}
\label{sec:introduction}
Large technology companies commonly maintain large relational datasets, derived from their internal logs, that can be represented as or joined into a massive heterogeneous graph (HG) composed of nodes and edges with multiple types~\cite{sun2012mining}.
For instance, in e-commerce networks, there are product, user, and review nodes, all interconnected by many edge types that represent forms of interactions such as spending (user-product), reviewing (user-review), and reviews-of (product-review).
To learn powerful features representing the complex multimodal structure of HGs, various heterogeneous graph neural networks (HGNN) have been proposed ~\cite{hu2020heterogeneous, schlichtkrull2018modeling, wang2019heterogeneous, zhang2019heterogeneous}.

A common issue in these industrial applications of HGNNs is the label imbalance among different node types. For instance, publicly available \emph{content} nodes -- such as those representing video, text, and image content -- are abundantly labelled, whereas labels for other types (such as \emph{user} or \emph{account} nodes) may be much more expensive to collect (or even not available, e.g.\ due to privacy restrictions).
This means that in most standard training settings, HGNN models can only learn to make good inferences for a few label-abundant node types, and can usually not make any inferences for the remaining node types, given the absence of any labels for them.

If there is a pair of \emph{label-abundant} and \emph{zero-labeled} node types which share an inference task, could we transfer knowledge between them?
One body of work has focused on transferring knowledge between nodes of the \emph{same} type from two \emph{different} HGs (i.e., graph-to-graph transfer learning)~\cite{huang2020hgt, yang2021domain}.
However, these approaches are not applicable in many real-world scenarios for three reasons.
First, any external large-scale HG that could be used in a graph-to-graph transfer learning setting would almost surely be proprietary. 
Second, even if practitioners could obtain access to an external industrial HG, it is unlikely the distribution of that (source) graph would match their target graph well enough to apply transfer learning.
Finally, node types suffering label scarcity are likely to suffer the same issue on other HGs (e.g.\ user nodes).

In this paper, we introduce a zero-shot transfer learning approach for a \emph{single} HG (assumed to be fully-owned by the practitioners), transferring knowledge from labelled to unlabelled node types. This setting is distinct from any graph-to-graph transfer learning scenarios, since the source and target domains exist in the same HG dataset, and are assumed to have different node types. Our model utilizes the shared context between source and target node types; for instance, in the e-commerce network, the latent (unknown) labels of user nodes can be strongly correlated with spending/reviewing patterns that are encoded in the cross-edges between user nodes and product/review nodes. We propose a novel zero-shot transfer learning problem for this HG learning setting as follows:
\begin{problem_inf}
\label{problem_definition}\emph{\textbf{Zero-shot cross-type transfer learning running on a HG:}}\\
    Given a heterogeneous graph $\mathcal{G}$ with node types $\{\mathbf{s}, \mathbf{t}, \cdots \}$ with abundant labels for source type $\mathbf{s}$ but no labels for target type $\mathbf{t}$, can we train HGNNs to infer the labels of target-type nodes?
\end{problem_inf}
\vspace{-2mm}
A na\"ive solution to this problem would be to re-use an HGNN pre-trained on the source nodes for target node inference, given that both domains exist in the same HG. However, as we show in our paper, HGNNs have distinct parameter sets for each node type~\cite{hu2020heterogeneous}, edge type~\cite{schlichtkrull2018modeling}, and meta-path type~\cite{fu2020magnn, wang2019heterogeneous}. 
These facts cause HGNNs to learn entirely different \emph{feature extractors} for nodes and edges of different types -- in other words, the final embeddings for source and target nodes are computed by different sets of parameters in HGNNs. Thus, a classifier pre-trained on source nodes will fail to perform well on inference tasks for target nodes. 
The field of domain adaptation (DA) targets this setting, seeking to transfer knowledge from a source domain with abundant labels to a target domain which lacks them~\cite{ganin2016domain, long2017conditional, long2017deep, shen2018wasserstein}.
However, distinct feature extractors across node types in HGNNs break a standard assumption of DA setting, namely that source and target domains share the same feature extractors (e.g., CNNs for both source and target image domains). As we demonstrate in this paper, in our problem setting, DA approaches fail to achieve the outstanding performance they are known for in computer vision and NLP.

In our work, we first dissect the gradient path of HGNN models to see how feature extractors are designed independently for each node type, and some empirical consequences.
Then we theoretically analyze how feature extractors across node types relate to each other and how their output distributions could be represented in terms of each other.  
We model this theoretical relationship between two feature extractors as a Knowledge Transfer Network (\method) which can be optimized to transform target embeddings to fit the source domain distribution.
We perform an extensive evaluation of our method on 18 different transfer learning tasks on HGs where we compare against state-of-the-art domain adaptation baselines.
Additionally, in order to understand which environments are ideal for transferring knowledge between different node types for HGs, we formulate a synthetic heterogeneous graph generator that allows us to study the sensitivity of these methods.

Our main contributions are:
\begin{itemize}[leftmargin=10pt,topsep=0pt,itemsep=-1ex,partopsep=1ex,parsep=1ex]
    \item {
		\textbf{Novel and practical problem definition:}
		To the best of our knowledge, \method is the first zero-shot cross-type transfer learning method running on a heterogeneous graph --- transfer knowledge across different node types within a heterogeneous graph.
	}
	\item {
		\textbf{Generality:}
		\method is a principled approach analytically induced from the architecture of HGNNs, thus applicable to any HGNN models, showing up to $960\%$ performance improvement for zero-labeled node inference across $6$ different HGNN models.
	}
	\item {
		\textbf{Effectiveness:}
		We show that \method outperforms state-of-the-art domain adaptation methods, being up to $73.3\%$ higher in MRR on $18$ different transfer learning tasks on HGs.
	}
	\item {
		\textbf{Sensitivity Analysis:}
		We provide a HG generator model to analyze how the node attribute and edge distributions of HGs affect the performance of \method and other methods on the task.
	}
\end{itemize}

\vspace{-2mm}
\section{Related Work}
\label{sec:related_work}
\vspace{-1mm}
\setlength{\textfloatsep}{2.5pt}
Various transfer learning problems have been defined on the graph domain.
\cite{luo2020progressive, ma2019gcan, wu2021towards, you2020handling} construct synthetic graphs from unstructured data and transfer knowledge over the graphs using GNNs.
On the other hand, \cite{hu2019strategies, hu2020gpt, qiu2020gcc, wu2020unsupervised} focus on extracting knowledge from the existing graph structures.
They pretrain a GNN model on a source graph and re-use the model on a target graph.
While these methods focus on homogeneous graphs, \cite{huang2020hgt, yang2021domain} transfer HGNNs across different HGs.
However, none of them can be directly applied to our cross-type transfer learning problem running on a single HG.
Here we cover two classes of learning approaches that are related to our problem.
As HGNNs are the models to which our method can be applied, we cover them extensively in Section \ref{sec:preliminaries}.

\vspace{-1mm}
\paragraph{Zero-shot domain adaptation (DA)} transfers knowledge from a source domain with abundant labels to a target domain which lacks them. Zero-shot DA can be categorized into three groups --- MMD-based methods, adversarial methods, and optimal-transport-based methods.
MMD-based methods~\cite{long2015learning, long2017deep,sun2016return} minimize the maximum mean discrepancy (MMD)~\cite{gretton2012kernel} between the mean embeddings of two distributions in reproducing kernel Hilbert space.
Adversarial methods~\cite{ganin2016domain, long2017conditional} are motivated by theory in~\cite{ben2010theory, ben2007analysis} suggesting that a good cross-domain representation contains no discriminative information about the origin of the input.
They learn domain-invariant features by a min-max game between the domain classifier and the feature extractor.
Optimal transport-based methods~\cite{shen2018wasserstein} estimate the empirical Wasserstein distance~\cite{redko2017theoretical} between two domains and minimizes the distance in an adversarial manner.
All three categories rely on two networks --- a feature extractor network and a task classifier network.
Adversarial and OT-based methods use an additional domain classifier network.
Due to the assumption that source and target domains have the same modality~\footnote{In our problem, source and target node types could have either (1) different distributions on the same attribute space or (2) entirely different attribute spaces}, the standard DA setting assumes identical feature extractors across domains.
More descriptions can be found in Appendix~\ref{appendix:baseline}.

\vspace{-1mm}
\paragraph{Label propagation (LP)} approaches (e.g.,~\cite{zhu2005semi}) use message-passing to pass each node's label to its neighbors according to normalized edge weights.
LP relies on only a graph's edges, and is therefore easily applied to a heterogeneous graph -- labels are simply propagated across edges, regardless of type.
In this paper we also evaluate a similarly-simple baseline, embedding propagation (EP).
Similar to LP, EP recursively propagates source embeddings (computed using source labels) until they reach the target domain.
EP exploits both node attribute information and the node adjacencies, but only uses the source node embeddings.


\vspace{-1mm}
\section{Preliminaries}
\label{sec:preliminaries}
\vspace{-1mm}
In this section we review heterogeneous graphs and heterogeneous graph neural networks (HGNNs). 
\vspace{-2mm}
\subsection{Heterogeneous graph}
\label{sec:preliminaries:hg}
\vspace{-1mm}
Heterogeneous graphs (HGs) are an important abstraction for modeling the relational data of multi-modal systems. 
Formally, a heterogeneous graph is defined as $\mathcal{G} = (\mathcal{V}, \mathcal{E}, \mathcal{T}, \mathcal{R})$ where
the node set $\mathcal{V}$; 
the edge set $\mathcal{E}$ consisting of ordered tuples $e_{ij}:=(i, j)$ with $i, j\in\mathcal{V}$, where $e_{ij}\in\mathcal{E}$ iff an edge exists from $i$ to $j$; 
the set of node types $\mathcal{T}$ with associated map $\tau:\mathcal{V}\mapsto\mathcal{T}$; the set of relation types $\mathcal{R}$ with associated map $\phi:\mathcal{E}\mapsto\mathcal{R}$.
This flexible formulation allows directed, multi-type edges. 
We additionally assume the existence of a node attribute vector $x_i\in\mathcal{X}_{\tau(i)}$ for each $i\in\mathcal{V}$, where $\mathcal{X}_{t}$ is an attribute matrix specific to nodes of type $t$ . 

\subsection{Heterogeneous Graph Neural Networks (HGNN)}
\label{sec:preliminaries:hgnn}

Various HGNN models have been proposed~\cite{hu2020heterogeneous, schlichtkrull2018modeling, wang2019heterogeneous, yang2021interpretable, zhang2019heterogeneous}.
Fully-specified HGNN models have specialized parameters for each node type~\cite{hu2020heterogeneous}, edge type~\cite{schlichtkrull2018modeling}, and meta-path type~\cite{fu2020magnn} to most effectively utilize the complex relationships encoded in the HG data structure.
In this paper, we use a flavor of HGNN known as a Heterogeneous Message-Passing Neural Network (HMPNN) as our base model on which to demonstrate \method (though \method can be implemented in almost any HGNN, as we show in experiments in Section~\ref{sec:experiments}). The HMPNN merely extends the standard MPNN~\cite{gilmer2017neural} by specializing all transformation and message matrices in each node/edge type. 
With its generality, HMPNN is itself a base model for RGCN~\cite{schlichtkrull2018modeling} and HGT~\cite{hu2020heterogeneous}, and is also widely used as a default HGNN model in popular GNN libraries (e.g., pyG~\cite{fey2019fast}, TF-GNN~\cite{ferludin2022tfgnn}, DGL~\cite{wang2019deep}).

In a HMPNN, for any node $j$, the embedding of node $j$ at the $l$-\emph{th} layer is obtained with the following generic formulation:
\begin{equation}\label{eq:gnn}
     \small
    h_j^{(l)} = \textbf{Transform}^{(l)}\left(\textbf{Aggregate}^{(l)}(\mathcal{E}(j))\right)
\end{equation}
where $\mathcal{E}(j) = \{(i, j)\in\mathcal{E}: i,j\in\mathcal{V}\}$ denotes all the edges which connect (directionally) to $j$.
The above operations typically involve type-specific parameters to exploit the inherent multiplicity of modalities in heterogeneous graphs.
First, we define a linear \textbf{Message} function:
\begin{equation}\label{eq:message}
    \small
    \textbf{Message}^{(l)}(i, j) = M_{\phi((i, j))}^{(l)}\cdot \left(h_i^{(l-1)}\concat h_j^{(l-1)}\right)
\end{equation}
where $M_r^{(l)}$ are the specific message passing parameters for each edge type $r\in\mathcal{R}$ and each of $L$ HMPNN layers. 
Then defining $\mathcal{E}_r(j)$ as the set of edges of type $r$ pointing to node $j$, the \textbf{Aggregate} function mean-pools messages by edge type, and concatenates:
\begin{equation}\label{eq:aggregate}
    \small
    \textbf{Aggregate}^{(l)}(\mathcal{E}(j)) = \underset{r \in\mathcal{R}}{\concat}\tfrac{1}{|\mathcal{E}_r(j)|}\sum_{e\in\mathcal{E}_r(j)}\textbf{Message}^{(l)}(e)
\end{equation}
Finally, the \textbf{Transform} function maps the message into a type-specific latent space:
\begin{equation}\label{eq:transform}
    \small
    \textbf{Transform}^{(l)}(j) = \alpha(W_{\tau(j)}^{(l)}\cdot\textbf{Aggregate}^{(l)}(\mathcal{E}(j)))
\end{equation}
where $W_t^{(l)}$ are the specific transformation parameters for each node type $t\in\mathcal{T}$ and each of $L$ HMPNN layers. 
By stacking $L$ layers, HMPNN outputs highly contextualized final node representations, and the final node representations can be fed into another model to perform downstream heterogeneous network tasks, such as node classification or link prediction.


\subsection{Problem definition}
\label{sec:preliminaries:problem_definition}

Using notations defined above, we formalize our novel transfer learning problem on HGs.
\begin{problem}
\label{problem_definition_formal}\emph{\textbf{Zero-shot cross-type transfer learning running on a HG:}}\\
    In a given heterogeneous graph $\mathcal{G} = (\mathcal{V}, \mathcal{E}, \mathcal{T}, \mathcal{R})$ with node attributes $\mathcal{X} = \cup_{t\in \mathcal{T}}\mathcal{X}_t$, assume node types $\mathbf{s}$ and $\mathbf{t}$ share a classification task $\{(i, y_i): i \in \mathcal{V}_s, \mathcal{V}_t\}$. 
    During the training phase, using labels $\{(i, y_i): i \in \mathcal{V}_s\}$ only for source-type nodes, we train an HGNN model $\textbf{f}: \textbf{f}(\mathcal{G}, \mathcal{X}) = h_i$ to get node embeddings $h_i$ for all nodes $i \in \mathcal{V}$ and a classifier $\textbf{g}: \textbf{g}(h_i) = \hat{y}_i$ to predict labels $\hat{y}_i$ from the node embeddings $h_i$. 
    During the test phase, our task is to predict labels $\{(j, y_j): j \in \mathcal{V}_t\}$ of target-type nodes where none of labels of target-type nodes were used for training.
\end{problem}

\section{Cross-Type Feature Extractor Transformations in HGNNs}
\label{sec:motivation}
We define $f_t:\mathcal{G}\mapsto\mathbb{R}^d$ to be the ``feature extractor" of a HGNN, which is composed of parameters participating to map input node attributes of type $t$ into a shared feature space $\mathbb{R}^d$.
In this section, we derive a strict transformation between feature extractors within a HMPNN. Specifically, for any two nodes $i,j$ with types $\tau(i) = s$ and $\tau(j) = t$, we derive an expression for $f_s$ in terms of $f_t$, and use that expression to inspire a trainable transfer learning module called \method in the following section.
For simplicity, throughout this section we ignore the activation $\alpha(\cdot)$ within the \textbf{Transform} function in Equation~\eqref{eq:transform}. 

\subsection{Feature extractors in HMPNNs}\label{sec:motivation:feature_extractor}
We first reason intuitively about the differences between $f_s$ and $f_t$ when $s\ne t$, using a toy heterogeneous graph shown in Figure~\ref{fig:toy:hg}. 
Consider nodes $v_1$ and $v_2$, noticing that $\tau(1)\ne \tau(2)$. 
Using HMPNN's equations~\eqref{eq:message}-\eqref{eq:transform} from Section \ref{sec:preliminaries:hgnn}, for any $l\in\{0, \ldots, L-1\}$ we have
\begin{align}\label{eq:tpynode1}
    \small
    h_1^{(l)} &= W_s^{(l)}\left[M_{ss}^{(l)}\left(h_3^{(l-1)}\concat h_1^{(l-1)}\right)\concat M_{ts}^{(l)}\left(h_2^{(l-1)}\concat h_1^{(l-1)}\right)\right] \\
    h_2^{(l)} &= W_t^{(l)}\left[M_{st}^{(l)}\left(h_1^{(l-1)}\concat h_2^{(l-1)}\right)\concat M_{tt}^{(l)}\left(h_4^{(l-1)}\concat h_2^{(l-1)}\right)\right]
\end{align}
where $h_j^{(0)} = x_j$. From these equations, we see that $h_1^{(l)}$ and $h_2^{(l)}$, which are features of different types, are extracted using \emph{disjoint} sets of model parameters at $l$-th layer. In a 2-layer HMPNN, this creates unique gradient backpropagation paths between the two node types, as illustrated in Figures~\ref{fig:toy:source_cg}-\ref{fig:toy:target_cg}. In other words, even though the same HMPNN is applied to node types $s$ and $t$, the feature extractors $f_s$ and $f_t$ have different computational paths. Therefore they project node features into different latent spaces, and have different update equations during training.

\subsection{Empirical gap between $f_s$ and $f_t$}
\label{sec:motivation:experiments}

\begin{figure*}[t!]
 	\centering
 	\vspace{-4mm}
 	\subfigure[Toy graph]
 	{
 	\label{fig:toy:hg}
 	\includegraphics[width=.15\linewidth]{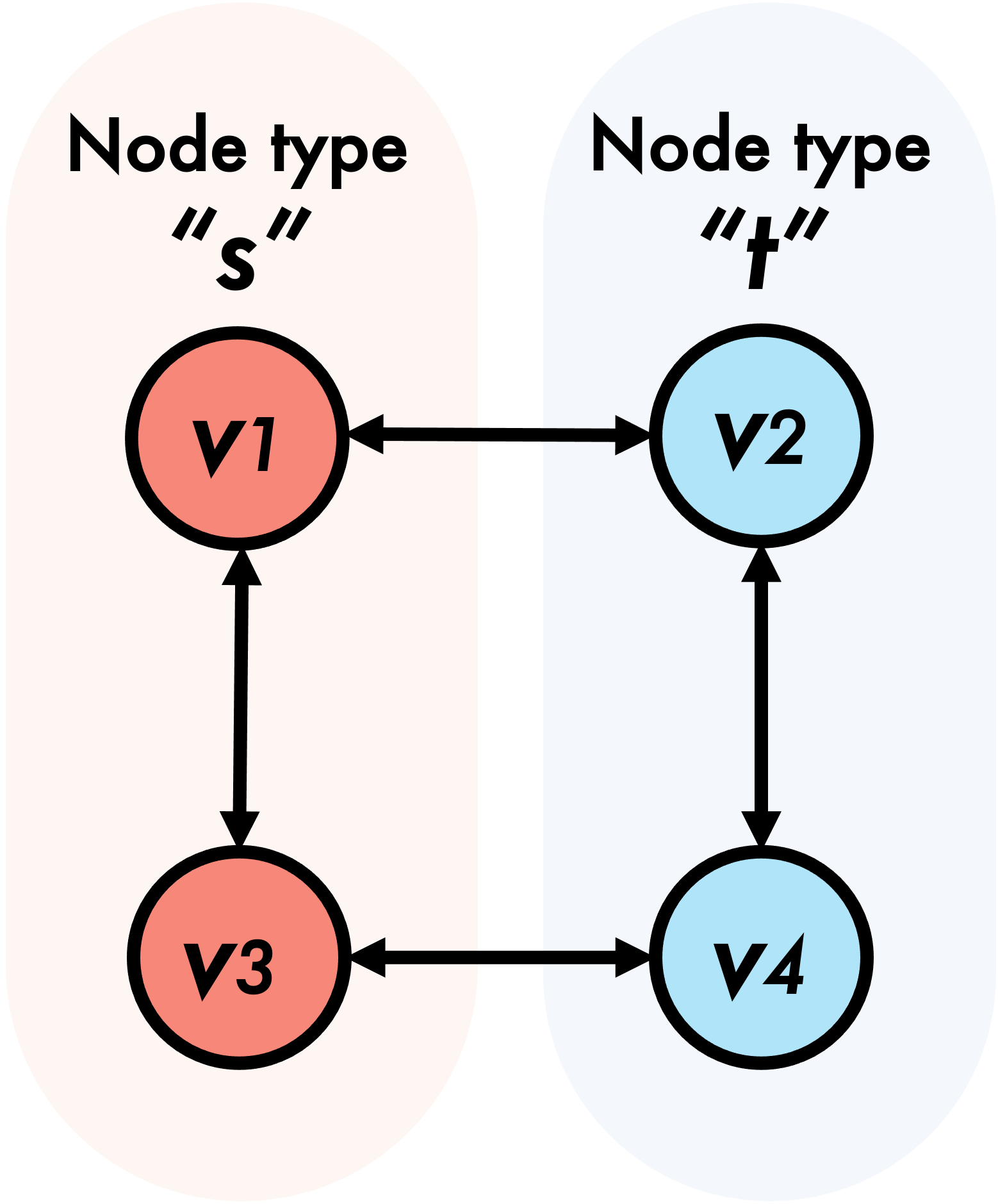}
 	}
 	\hspace{7mm}
 	\subfigure[Gradient path for feature extractor $f_s$]
 	{
 	\label{fig:toy:source_cg}
 	\includegraphics[width=.25\linewidth]{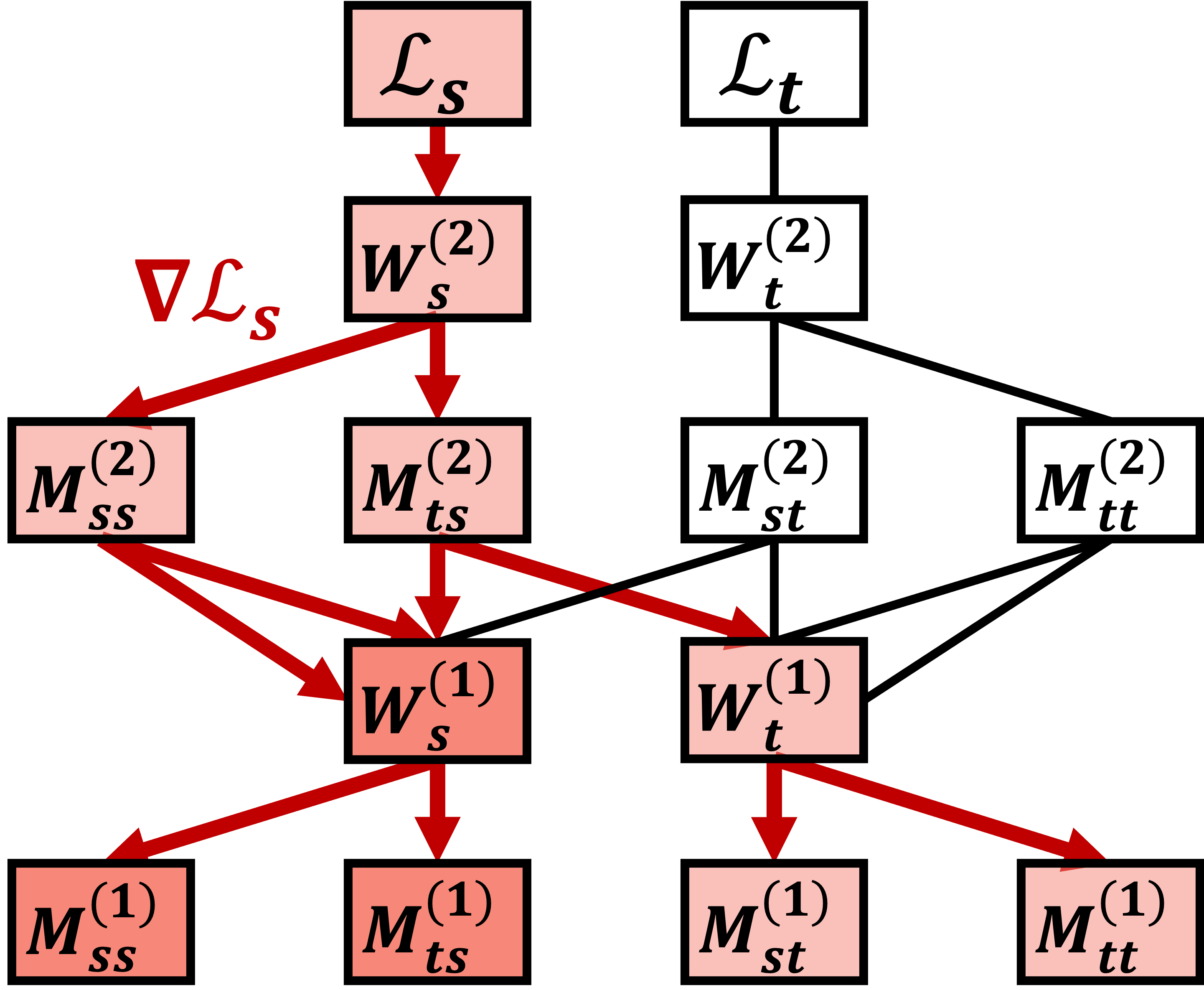}
 	}
 	\hspace{7mm}
 	\subfigure[Gradient path for feature extractor $f_t$]
 	{
 	\label{fig:toy:target_cg}
 	\includegraphics[width=.24\linewidth]{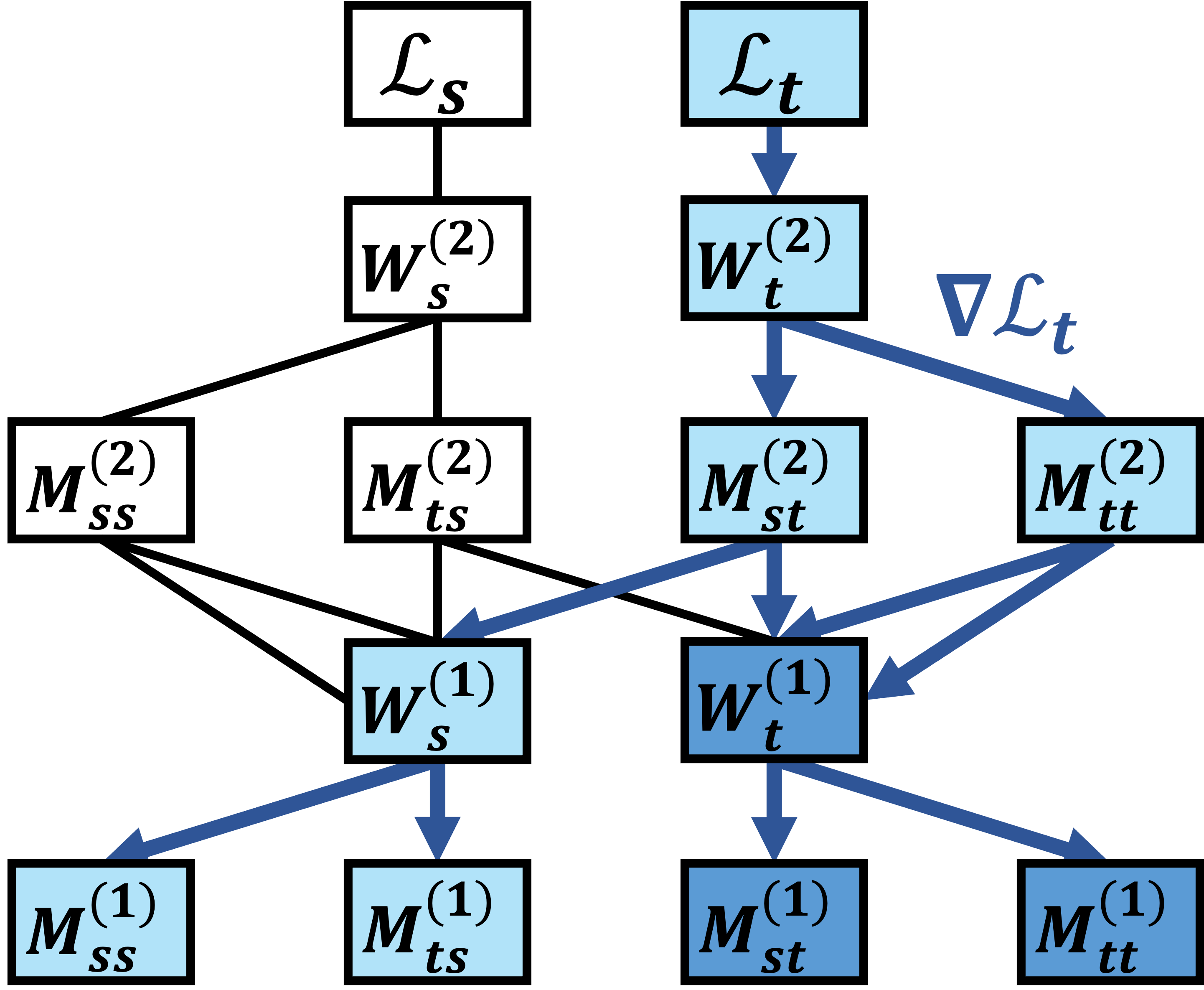}
 	}
 	\caption
 	{ \small
 	    Illustration of a toy heterogeneous graph and the gradient paths for feature extractors $f_s$ and $f_t$. 
 	    Colored arrows in figures (b) and (c) show that the same HGNN nonetheless produces different gradient paths for each feature extractor.
 	    Color density of each box in (b) and (c) is proportional to the degree of participation of the corresponding parameter in each feature extractor.
 	}
 	\label{fig:toy}
 \end{figure*}

Here we study the experimental consequences of the above observation via simulation. 
We first construct a synthetic graph extending the 2-type graph in Figure~\ref{fig:toy:hg} to have multiple nodes per-type, and multiple classes. 
To maximize the effects of having different feature extractors, we sample source and target nodes from the same feature distributions and each classes are well-separated in the both the graph and feature space (more details available in Appendix~\ref{appendix:graph-generator:toy}). 

\begin{figure*}[t!]
 	 \vspace{-4mm}
 	\centering
 	\subfigure[Test accuracy across various feature extractors]
 	{
 	\label{fig:toy_exp:accuracy}
 	\includegraphics[width=0.32\linewidth]{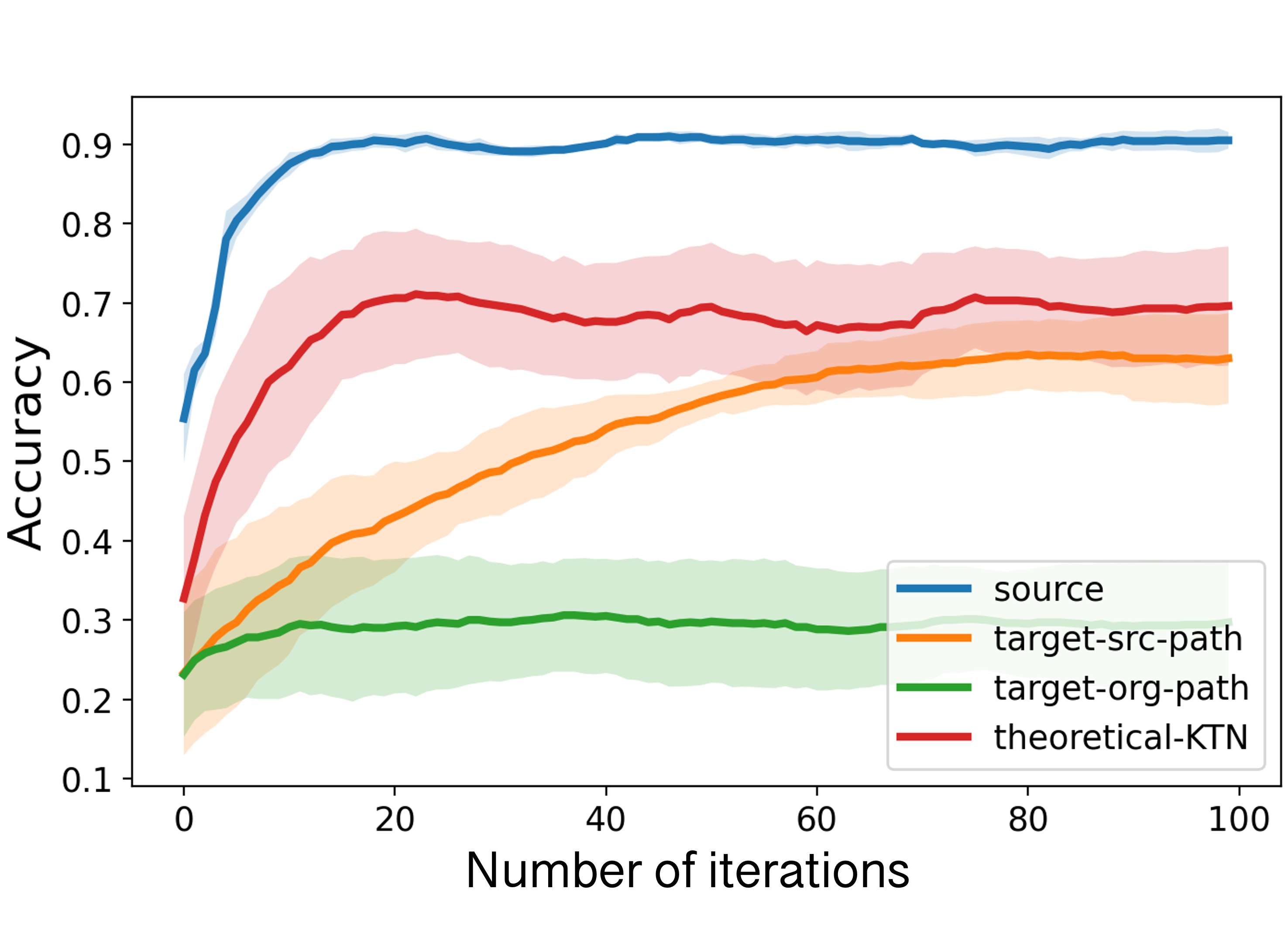}
 	}
 	\hspace{3mm}
 	\subfigure[L2 norms of gradients of $W_{\tau(\cdot)}$]
 	{
 	\label{fig:toy_exp:transform_w}
 	\includegraphics[width=.28\linewidth]{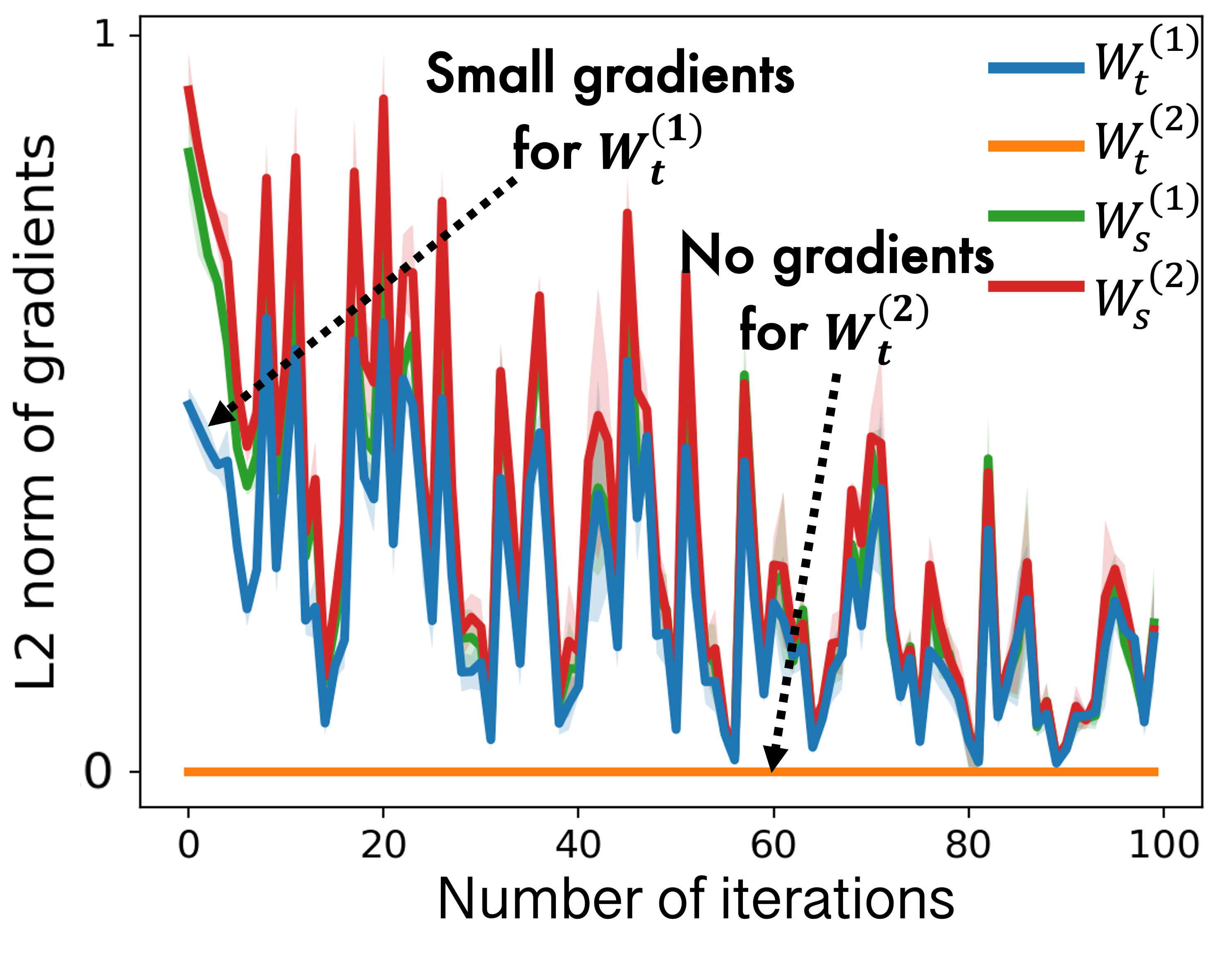}
 	}
 	\hspace{3mm}
 	\subfigure[L2 norms of gradients of $M_{\phi(\cdot)}$]
 	{
 	\label{fig:toy_exp:message_w}
 	\includegraphics[width=.28\linewidth]{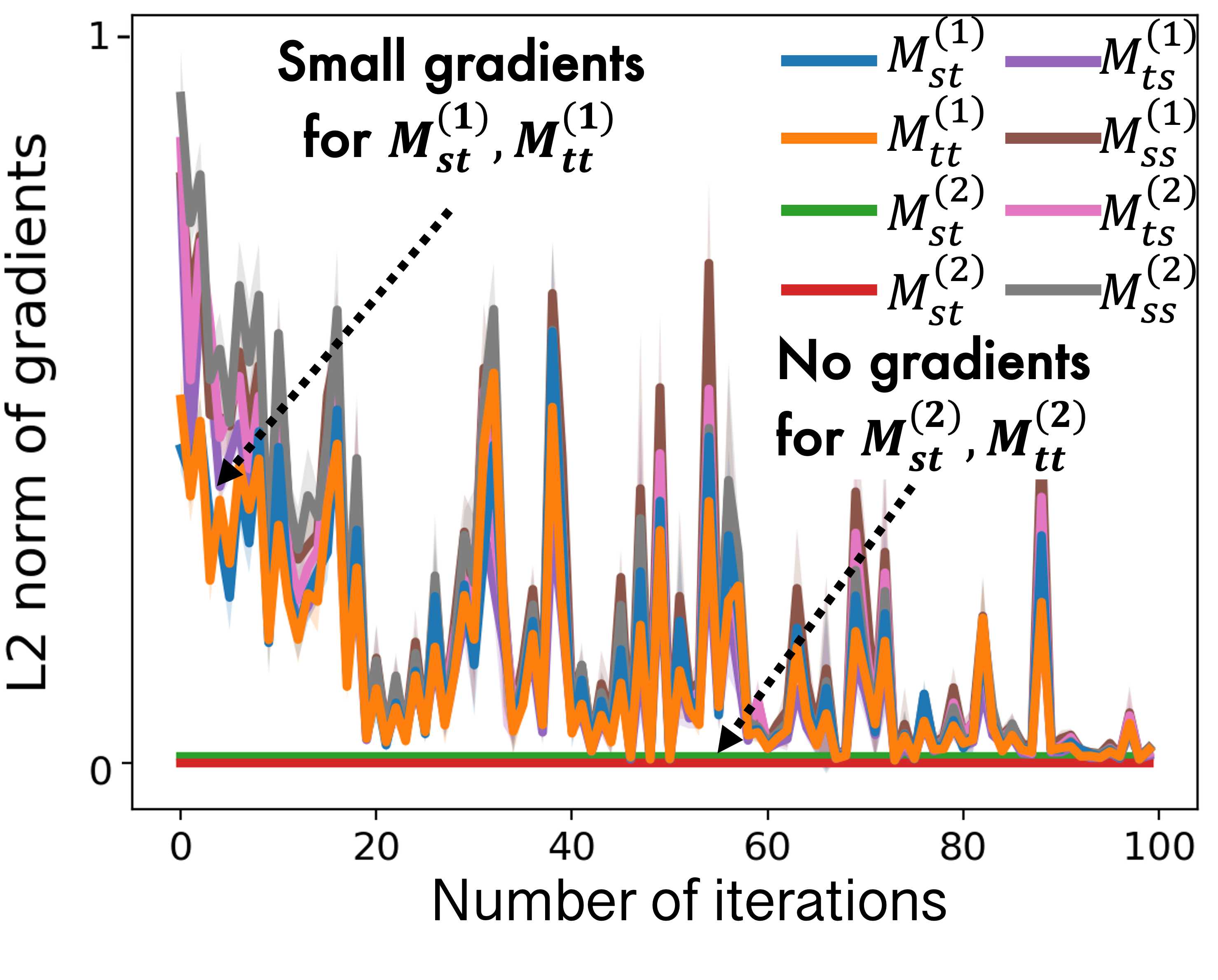}
 	}
 	\caption
 	{ \small
 	    HGNNs trained on a source domain underfit a target domain even on a ``nice" heterogeneous graph.   	     
 	    (a) Performance on the simulated heterogeneous graph for 4 kinds of feature extractors; (\textit{source}: source extractor $f_s$ on source domain,
 	    \textit{target-src-path}: source extractor $f_s$ on target domain, 
 	    \textit{target-org-path}: target extractor $f_t$ on target domain, 
 	    and \textit{theoretical-KTN}: target extractor $f_t$ on target domain using \method.)
 	    (b-c) L2 norms of gradients of parameters $W_{\tau(\cdot)}$ and $M_{\phi(\cdot)}$ in HGNN models.
 	}
 	\vspace{3mm}
 	\label{fig:toy_exp}
\end{figure*}

On such a well-aligned heterogeneous graph, without considering the observation in Section~\ref{sec:motivation:feature_extractor}, there may seem to be no need for domain adaptation from $f_t$ to $f_s$. 
However, when we train the HMPNN model solely on $s$-type nodes, as shown in Figure~\ref{fig:toy_exp:accuracy} we find the test accuracy for $s$-type nodes to be high ($90\%$, blue line) and the test accuracy for $t$-type nodes to be quite low ($25\%$, green line). 
Now if instead we make the $t$-type nodes use the source feature extractor $f_s$, much more transfer learning is possible (${\sim}65\%$, orange line). 
This shows that the different feature extractors present in the HMPNN model result in the significant performance drop, and simply matching input data distributions can not solve the problem.

To analyze this phenomenon at the level of backpropagation, in Figures~\ref{fig:toy_exp:transform_w}-\ref{fig:toy_exp:message_w} we show the magnitude of gradients passed to parameters of source and target node types. As illustrated in Figures~\ref{fig:toy:source_cg}-\ref{fig:toy:target_cg}, we find that the final-layer \textbf{Transform} parameter $W^{(2)}_t$ for type-$t$ nodes have zero gradients (Figure~\ref{fig:toy_exp:transform_w}), and similarly for the final-layer \textbf{Message} parameters (Figure~\ref{fig:toy_exp:message_w}). Additionally, those same parameters in the first-layer for $t$-type nodes have much smaller gradients than their $s$-type counterparts: $W^{(1)}_{t}$ (blue line in Figure~\ref{fig:toy_exp:transform_w}), $M^{(1)}_{st}$ and $M^{(1)}_{tt}$ (blue and orange lines in Figure~\ref{fig:toy_exp:message_w}) appear below than other lines. This is because they contribute to $f_s$ less than $f_t$

This case study shows that even when an HGNN is trained on a relatively simple, balanced, and class-separated heterogeneous graph, a model trained only on the source domain node type cannot transfer to the target domain node type.

\subsection{Relationship between feature extractors in HMPNNs}
\label{sec:motivation:theoretical_analysis}

We show that a HMPNN model provides different feature extractors for each node type. 
However, still, $f_s$ and $f_t$ are built inside one HMPNN model and interchange intermediate feature embeddings with each other.
Both $H_s^{(L)}$ and $H_t^{(L)}$ (the output of $f_s$ and $f_t$) are computed using the previous layer's intermediate embeddings $H_s^{(L-1)}, H_t^{(L-1)}$, and any other connected node type embeddings $H_x^{(L-1)}$ at the $L$-th HMPNN layer.
Therefore $H_s^{(L)}$ and $H_t^{(L)}$ can be mathematically presented by each other using the $(L-1)$-th layer embeddings as connecting points, so do $f_s$ and $f_t$.
Based on this intuition, we derive a strict transformation between $f_s$ and $f_t$, which will motivate the core domain adaptation component of our proposed \method model.

\begin{theorem}\label{theorem}
Given a heterogeneous graph $\mathcal{G} = \{\mathcal{V}, \mathcal{E}, \mathcal{T}, \mathcal{R}\}$. For any layer $l>0$, define the set of $(l-1)$-\emph{th} layer HMPNN parameters as
\begin{equation}
\label{eq:thm}
    \small
    \mathcal{Q}^{(l-1)} = \{M_r^{(l-1)}: r\in\mathcal{R}\}\cup\{W_t^{(l-1)}: t\in\mathcal{T}\}.
\end{equation}

Let $A$ be the total $n\times n$ adjacency matrix. Then for any $s,t\in\mathcal{T}$ there exist matrices $A_{ts}^\ast = a_{ts}(A)$ and $Q_{ts}^\ast = q_{ts}(\mathcal{Q}^{(l-1)})$ such that
\begin{equation}
\label{eq:relationship}
    \small
    H_s^{(l)} = A_{ts}^\ast H_t^{(l)} Q_{ts}^\ast
\end{equation}
where $a_{ts}(\cdot)$ and $q_{ts}(\cdot)$ are matrix functions that depend only on $s,t$.
\end{theorem}
\vspace{2mm}
The full proof of Theorem 1 can be found in Appendix~\ref{appendix:theorem1}.  Notice that in Equation~\ref{eq:relationship}, $Q_{ts}^{\ast}$ acts as a macro-$\textbf{Transform}$ operator that maps $H^{(L)}_t$ into the source domain, then $A_{ts}^{\ast}$ aggregates the mapped embeddings into $s$-type nodes. 
In other words, $Q_{ts}^{\ast}$ acts as a mapping matrix from the target domain to the source domain.
To examine the implications, we run the same experiment as described in Section~\ref{sec:motivation:experiments}, while this time mapping the target features $H^{(L)}_t$ into the source domain by multiplying with $Q_{ts}^{\ast}$ in Equation~\ref{eq:relationship} before passing over to a task classifier.
We see via the red line in Figure~\ref{fig:toy_exp:accuracy} that, with this mapping, the accuracy in the target domain becomes much closer to the accuracy in the source domain (${\sim}70\%$).  Thus, we use this theoretical transformation as a foundation for our trainable HGNN domain adaptation module, introduced in the following section.

\subsection{Generalized cross-type transformations for HGNNs}
\label{sec:motivation:generalizability}
In this section we showed that a HMPNN feature extractor on the (label-abundant) source node type can be expressed in terms of the (label-scarce) target node type feature extractor, and this transformation enables cross-type zero-shot learning for the target node type. 
As most HGNNs have distinct feature extractors for each node types (even single-layer HGNNs, which have specialized parameters for each node/edge attribute projection layer), they will suffer from the under-trained target embeddings phenomena we showed in Section~\ref{sec:motivation:experiments}.
For instance, in the meta-path based MAGNN model~\cite{fu2020magnn}, meta-paths directing toward the target node types are generally less engaged in the source node feature computation and thus receive smaller gradients. 
While we cannot derive the exact cross-type transformation for all possible HGNNs, 
the core intuition in the HMPNN theory holds, namely that $H_s^{(L)}$ and $H_t^{(L)}$ are both computed using the previous layer's intermediate embeddings (see Section~\ref{sec:motivation:theoretical_analysis}) across all HGNN models. 
This observation allows us to extend our \method and apply it to almost any HGNN.
We illustrate this by applying \method to $6$ different HGNN models in Section~\ref{sec:experiments}, where we see greatly increased target-type accuracy.

\section{\method: Trainable Cross-Type Transfer Learning for HGNNs}
\label{sec:matching_loss}
\begin{algorithm}[t!]
    \caption{Training step on a source domain}
    \label{alg:train}
\begin{algorithmic}[1]
\small
    \REQUIRE heterogeneous graph $\mathcal{G} = (\mathcal{V}, \mathcal{E}, \mathcal{T}, \mathcal{R})$, node feature matrices $\mathcal{X}$, source node type $s$, target node type $t$, adjacency matrix $A_{ts}$, source node label matrix $\mathcal{Y}_s$.
    \ENSURE HGNN $\textbf{f}$, classifier $\textbf{g}$, \method $\textbf{t}_{\text{KTN}}$
    \STATE $H^{(L)}_s, H^{(L)}_t = \textbf{f}(\mathcal{G}, H^{(0)} = \mathcal{X})$
    \STATE $H^{*}_{t} = \textbf{t}_{KTN}(H^{(L)}_t) = A_{ts}H^{(L)}_tT_{ts}$ 
    \STATE $\mathcal{L}_{\text{KTN}} = \left\|H^{(L)}_{s} - H^{*}_{t}\right\|_{2}$
    \STATE $\mathcal{L} = \mathcal{L}_{\text{CL}}(\textbf{g}(H^{(L)}_s), \mathcal{Y}_s) + \lambda\mathcal{L}_{\text{KTN}}$
    \STATE Update $\textbf{f}$, $\textbf{g}$, $\textbf{t}$ using $\nabla\mathcal{L}$
\normalsize
\end{algorithmic}
\end{algorithm}

\begin{algorithm}[t!]
    \caption{Test step on a target domain}
    \label{alg:test}
\begin{algorithmic}[1]
\small
    \REQUIRE pretrained HGNN $\textbf{f}$, classifier $\textbf{g}$, \method $\textbf{t}_{\text{KTN}}$
    \ENSURE target node label matrix $\mathcal{Y}_t$
    \STATE $H^{(L)}_t = \textbf{f}(\mathcal{G}, H^{(0)} = \mathcal{X})$ 
    \STATE $H^{*}_{t} = H^{(L)}_tT_{ts}$
    \RETURN $\textbf{g}(H^{*}_{t})$
\normalsize
\end{algorithmic}
\end{algorithm}

Inspired by these derivations we introduce our primary contribution, \textit{Knowledge Transfer Networks}.
We begin by noting Equation~\ref{eq:relationship} in Theorem~\ref{theorem} has a similar form to a single-layer graph convolutional network~\citep{kipf2016semi} with a deterministic transformation matrix ($Q_{ts}^{\ast}$) and a combination of adjacency matrices directing from target node type $t$ to source node type $s$ ($A_{ts}^\ast$).
Instead of hand-computing the mapping function $Q_{ts}^{\ast}$ for arbitrary HGs and HGNNs (which would be intractable), we \emph{learn} the mapping function by modelling Equation~\ref{eq:relationship} as a trainable graph convolutional network, named the Knowledge Transfer Network, $\textbf{t}_{\text{\method}}(\cdot)$.
\method replaces $Q_{ts}^{\ast}$ and $A_{ts}^\ast$ in Equation~\ref{eq:relationship} as follows:
\begin{align}
    \small
    &\textbf{t}_{\text{KTN}}(H^{(L)}_{t}) = A_{ts} H^{(L)}_{t} T_{ts} \\
    &\mathcal{L}_{\text{KTN}} = \left\|H^{(L)}_{s} - \textbf{t}_{\text{KTN}}(H^{(L)}_{t})\right\|_{2}
\end{align}
\noindent
where $A_{ts}$ is an adjacency matrix from node type $t$ to $s$, and $T_{ts}$ is a trainable transformation matrix.
By minimizing $\mathcal{L}_{\text{KTN}}$, $T_{ts}$ is optimized to a mapping function of the target domain into the source domain.

\subsection{Algorithm}
\label{sec:matching_loss:algorithm}

We minimize a classification loss $\mathcal{L}_{\text{CL}}$ and a transfer loss $\mathcal{L}_{\text{KTN}}$ jointly with regard to a HGNN model $\textbf{f}$, a classifier $\textbf{g}$, and a knowledge transfer network $\textbf{t}_{\text{KTN}}$ as follows:
\begin{align*}
    \small
    \underset{\textbf{f},~\textbf{g},~\textbf{t}_{\text{KTN}}}{min}\mathcal{L}_{\text{CL}}(\textbf{g}(\textbf{f}(\mathcal{G}, \mathcal{X})_{s}), \mathcal{Y}_{s}) + \lambda \left\|\textbf{f}(\mathcal{G}, \mathcal{X})_{s} - \textbf{t}_{\text{KTN}}(\textbf{f}(\mathcal{G}, \mathcal{X})_{t})\right\|_{2}
\end{align*}
where $\lambda$ is a hyperparameter regulating the effect of  $\mathcal{L}_{\text{KTN}}$; and $\textbf{f}(\mathcal{G}, \mathcal{X})_{s}$ and $\textbf{f}(\mathcal{G}, \mathcal{X})_{t}$ denote $H^{(L)}_s$ and $H^{(L)}_t$, respectively.
Algorithm~\ref{alg:train} describes a training step on the source domain.
After computing the node embeddings $H^{(L)}_s$ and $H^{(L)}_t$, we map $H^{(L)}_t$ to the source domain using $\textbf{t}_{\text{KTN}}$ and compute $\mathcal{L}_{\text{KTN}}$.
Then, we update the models using gradients of $\mathcal{L}_{\text{CL}}$ (computed using only source labels) and $\mathcal{L}_{\text{KTN}}$.
Algorithm~\ref{alg:test} describes the test phase on the target domain.
After we get node embeddings $H^{(L)}_t$ from the trained HGNN model, we map $H^{(L)}_t$ into the source domain using the trained transformation matrix $T_{ts}$.
Finally we pass the transformed target embeddings $H^{*}_t$ into the classifier which was trained on the source domain.

\noindent \textbf{Indirect Connections}
We note that in practice, the source and target node types can be indirectly connected in HGs via other node types (i.e., there is no $A_{ts}$).
Appendix~\ref{appendix:indirect} describes how we can easily extend \method to cover domain adaption scenarios in this case.

\section{Experiments}
\label{sec:experiments}
\subsection{Datasets}
\label{sec:experiments:dataset}

\noindent \textbf{Open Academic Graph (OAG).}~ A dataset introduced in  \cite{zhang2019oag} composed of five types of nodes: papers (P), authors (A), institutions (I), venues (V), fields (F) and their corresponding relationships.
Paper and author nodes have text-based attributes, while institution, venue, and field nodes have text- and graph structure-based attributes.
Paper, author, and venue nodes are labeled with research fields in two hierarchical levels, L1 and L2.
We construct three field-specific subgraphs from OAG: computer science, computer networks, and machine learning academic graphs. \vspace{-5pt}

\noindent \textbf{PubMed.}\citep{yang2020heterogeneous}
A network composed of of four types of nodes: genes (G), diseases (D), chemicals (C), and species (S), and their corresponding relationships.
Gene and chemical nodes have graph structure-based attributes, while disease and species nodes have text-based attributes.
Each gene and disease is labeled with a set of diseases they belong to or cause. \vspace{-3pt}

\noindent \textbf{Synthetic heterogeneous graphs.} We generate stochastic block models \citep{abbe2017community} with multiple node/edge types. 
We label each node types with the same set of classes.
Then we control feature/edge distributions within/between node types by manipulating feature/edge signal-to-noise ratio within/between classes.
A complete definition of the generative model is given in Appendix~\ref{appendix:graph-generator}.

\subsection{Baselines}
\label{sec:experiments:baseline}

We compare \method with two MMD-based DA methods (DAN~\cite{long2015learning}, JAN~\cite{long2017deep}), three adversarial DA methods (DANN~\cite{ganin2016domain}, CDAN~\cite{long2017conditional}, CDAN-E~\cite{long2017conditional}), one optimal transport-based method (WDGRL~\cite{shen2018wasserstein}), and two traditional graph mining methods (LP and EP~\cite{zhu2005semi}).
For \method and DA methods, we use HMPNN as their base HGNN model.
More information of each method is in Appendix~\ref{appendix:baseline}.

\begin{table*}[t]
    \caption{
    \small
	\textbf{Open Academic Graph on Computer Science field}.  The \textit{gain} column shows the relative gain of our method over using no domain adaptation (\textit{Base} column).
	\textit{o.o.m} denotes \textit{out-of-memory} errors.
	}
	\label{tab:oag:cs}
	\centering
    \tiny
\begin{tabular}{l|l|c|cc|ccc|c|cc|r}
\toprule\hline
\textbf{Task}                      & \textbf{Metric} & \textbf{Base} & \textbf{DAN} & \textbf{JAN} & \textbf{DANN} & \textbf{CDAN} & \textbf{CDAN-E} & \textbf{WDGRL} & \textbf{LP} & \textbf{EP} & \textbf{KTN (gain)}  \\ \hline\midrule
\multirow{2}{*}{\textbf{P-A (L1)}} & \textbf{NDCG}   & 0.399           & 0.452        & 0.405        & 0.292         & 0.262         & 0.261           & 0.260           & 0.178       & 0.425       & \textbf{0.623 (56$\%$)}  \\
                                   & \textbf{MRR}    & 0.297          & 0.361        & 0.314        & 0.179         & 0.129         & 0.111           & 0.138          & 0.041       & 0.363       & \textbf{0.629 (112$\%$)} \\ \hline
\multirow{2}{*}{\textbf{A-P (L1)}} & \textbf{NDCG}   & 0.401          & 0.566        & 0.598        & 0.294         & 0.364         & 0.246           & 0.195          & 0.153       & 0.557       & \textbf{0.733 (83$\%$)}  \\
                                   & \textbf{MRR}    & 0.318          & 0.508        & 0.544        & 0.229         & 0.270          & 0.090            & 0.047          & 0.022       & 0.507       & \textbf{0.711 (123$\%$)} \\ \hline
\multirow{2}{*}{\textbf{A-V (L1)}} & \textbf{NDCG}   & 0.459          & 0.457        & 0.470         & 0.382         & 0.346         & 0.359           & 0.403          & 0.207       & 0.461       & \textbf{0.671 (46$\%$)}  \\
                                   & \textbf{MRR}    & 0.364          & 0.413        & 0.458        & 0.341         & 0.205         & 0.253           & 0.327          & 0.011       & 0.389       & \textbf{0.698 (92$\%$)}  \\ \hline
\multirow{2}{*}{\textbf{V-A (L1)}} & \textbf{NDCG}   & 0.283           & 0.443        & 0.435        & 0.242         & 0.372         & 0.418           & 0.272          & 0.153       & 0.154       & \textbf{0.584 (107$\%$)} \\
                                   & \textbf{MRR}    & 0.133          & 0.365        & 0.345        & 0.094         & 0.241         & 0.444           & 0.144          & 0.006       & 0.006       & \textbf{0.586 (340$\%$)} \\ \hline
\multirow{2}{*}{\textbf{P-A (L2)}} & \textbf{NDCG}   & 0.229          & 0.230         & o.o.m        & 0.239         & o.o.m         & o.o.m           & 0.168          & o.o.m       & 0.215       & \textbf{0.282 (23$\%$)}  \\
                                   & \textbf{MRR}    & 0.121           & 0.118        & o.o.m        & 0.140          & o.o.m         & o.o.m           & 0.020           & o.o.m       & 0.143       & \textbf{0.2248 (86$\%$)} \\ \hline
\multirow{2}{*}{\textbf{A-P (L2)}} & \textbf{NDCG}   & 0.197          & 0.162        & o.o.m        & 0.204         & 0.158         & 0.161           & 0.132          & o.o.m       & 0.208       & \textbf{0.287 (46$\%$)}  \\
                                   & \textbf{MRR}    & 0.095           & 0.052        & o.o.m        & 0.106         & 0.032         & 0.045           & 0.017          & o.o.m       & 0.132       & \textbf{0.242 (155$\%$)} \\ \hline
\multirow{2}{*}{\textbf{A-V (L2)}} & \textbf{NDCG}   & 0.347           & 0.329        & 0.295        & 0.325         & 0.288         & 0.273           & 0.289          & o.o.m       & 0.297       & \textbf{0.402 (16$\%$)}  \\
                                   & \textbf{MRR}    & 0.310           & 0.296        & 0.198        & 0.223         & 0.128         & 0.097           & 0.110           & o.o.m       & 0.227       & \textbf{0.399 (29$\%$)}  \\ \hline
\multirow{2}{*}{\textbf{V-A (L2)}} & \textbf{NDCG}   & 0.235          & 0.249        & 0.251        & 0.214         & 0.197         & 0.205           & 0.217          & o.o.m       & 0.119       & \textbf{0.252 (7$\%$)}   \\
                                   & \textbf{MRR}    & 0.129          & 0.157        & 0.161        & 0.090          & 0.044         & 0.068           & 0.085          & o.o.m       & 0.000           & \textbf{0.166 (28$\%$)} \\ \hline \bottomrule
\end{tabular}
\normalsize
\end{table*}

\begin{table*}[]
    \caption{
    \small
	\textbf{PubMed graph}. 
	The \textit{gain} column shows the relative gain over using \textit{Base} column.
	}
	\label{tab:pubmed}
	\centering
    \tiny
\begin{tabular}{l|l|c|cc|ccc|c|cc|r}
\toprule\hline
\textbf{Task}                 & \textbf{Metric} & \textbf{Base} & \textbf{DAN} & \textbf{JAN} & \textbf{DANN} & \textbf{CDAN} & \textbf{CDAN-E} & \textbf{WDGRL} & \textbf{LP} & \textbf{EP} & \textbf{KTN (gain)}        \\ \hline\midrule
\multirow{2}{*}{\textbf{D-G}} & \textbf{NDCG}   & 0.587           & 0.629        & 0.615        & 0.614         & 0.624         & 0.646           & 0.604          & 0.601       & 0.571       & \textbf{0.700 (19$\%$)} \\
                              & \textbf{MRR}    & 0.372           & 0.425        & 0.414        & 0.397         & 0.428         & 0.443           & 0.388          & 0.389       & 0.336       & \textbf{0.499 (34$\%$)} \\ \hline
\multirow{2}{*}{\textbf{G-D}} & \textbf{NDCG}   & 0.596           & 0.599        & 0.577        & 0.599         & 0.581         & 0.606           & 0.578          & 0.576       & 0.580       & \textbf{0.662 (11$\%$)} \\
                              & \textbf{MRR}    & 0.354           & 0.362        & 0.332        & 0.356         & 0.337         & 0.362           & 0.340          & 0.351       & 0.353       & \textbf{0.445 (26$\%$)} \\ \hline\bottomrule
\end{tabular}
\normalsize
\vspace{3mm}
\end{table*}

\subsection{Zero-shot transfer learning}
\label{sec:experiments:zero-shot}

We run $18$ different zero-shot transfer learning tasks across three OAG and PubMed graphs.
We run each experiment $3$ times and report the average value.
Due to the space limitation, we report the standard deviations and results on OAG-computer networks and OAG-machine learning in Appendix~\ref{appendix:result:da}.
Each heterogeneous graph has the same node classification task for both source and target node types.
During training, we are given 1) the heterogeneous graph structure information (i.e., adjacency matrices), 2) input node attribute matrices for all node types, and 3) labels on source-type nodes for the classification task.
During the test phase, we predict labels on target-type nodes for the same classification task.
The performance is evaluated by NDCG and MRR --- widely adopted ranking metrics~\cite{hu2020gpt, hu2020heterogeneous}.

In Tables~\ref{tab:oag:cs} and ~\ref{tab:pubmed}, our proposed method \method consistently outperforms all baselines on all tasks and graphs by up to $73.3\%$ higher in MRR (P-A(L1) task in OAG-CS, Table \ref{tab:oag:cs}).
When we compare with the base accuracy using the model pretrained on the source domain without any domain adaptation ($3$rd column, \textit{Base}), the results are even more impressive.
We see our method \method provides relative gains of up to $340\%$ higher MRR without using any labels from the target domain.
These results show the clear effectiveness of \method on zero-shot transfer learning tasks on a heterogeneous graph.
We mention that venue and author node types are not directly connected in the OAG graphs (Figure~\ref{fig:schema1:oag} in Appendix), but \method successfully transfer knowledge by passing intermediate node types. 


\noindent \textbf{Baseline Performance.}
\label{sec:experiments:zero-shot-analysis}
Among baselines, MMD-based models (DAN and JAN) outperform adversarial-based methods (DANN, CDAN, and CDAN-E) and optimal transport-based method (WDGRL), unlike results reported in~\cite{long2017conditional, shen2018wasserstein}.
These reversed results are a consequence of HGNN's unique feature extractors for each domains.
Adversarial- and optimal transport-based methods define separate losses for source and target feature extractors (which are not separated in their shared feature extractor assumption), resulting in divergent gradients between different feature extractors and poor domain adaption performance.
This shows again the importance of considering different feature extractors in HGNNs.
More analysis can be found in Appendix~\ref{appendix:analysis}.

\begin{table*}[]
    \caption{
    \small
	\textbf{\method on different HGNN models.} 
	The \textit{Source} column shows accuracy on for source node types. 
	\textit{Base} and \textit{KTN} columns show accuracy for target node types without/with using \method, respectively.
	The \textit{Gain} column shows the relative gain of our method over using no domain adaptation.
	}
	\label{tab:hgnn}
	\centering
    \tiny
\begin{tabular}{l|l|cccr|cccr} \toprule\hline
                   &                 & \multicolumn{4}{c}{\textbf{P-A (L1)}}                            & \multicolumn{4}{c}{\textbf{A-P (L1)}}                            \\
\textbf{HGNN type} & \textbf{Metric} & \textbf{Source} & \textbf{Base} & \textbf{KTN} & \textbf{Gain}   & \textbf{Source} & \textbf{Base} & \textbf{KTN} & \textbf{Gain}   \\ \midrule\hline
\textbf{R-GCN}     & \textbf{NDCG}   & 0.765           & 0.337        & 0.577        & \textbf{71.12\%}  & 0.648           & 0.388         & 0.647        & \textbf{66.82\%}  \\
\textbf{}          & \textbf{MRR}    & 0.757           & 0.236        & 0.587        & \textbf{148.73\%} & 0.623           & 0.270         & 0.611        & \textbf{126.18\%} \\\hline
\textbf{HAN}       & \textbf{NDCG}   & 0.476           & 0.179        & 0.520        & \textbf{190.56\%} & 0.515           & 0.182         & 0.512        & \textbf{181.33\%} \\
\textbf{}          & \textbf{MRR}    & 0.416           & 0.047        & 0.497        & \textbf{960.55\%} & 0.509           & 0.055         & 0.527        & \textbf{850.90\%} \\\hline
\textbf{HGT}       & \textbf{NDCG}   & 0.757           & 0.294        & 0.574        & \textbf{95.07\%}  & 0.670           & 0.283         & 0.581        & \textbf{104.83\%} \\
\textbf{}          & \textbf{MRR}    & 0.749           & 0.178        & 0.563        & \textbf{216.17\%} & 0.670           & 0.149         & 0.565        & \textbf{279.52\%} \\\hline
\textbf{MAGNN}     & \textbf{NDCG}   & 0.657           & 0.463        & 0.574        & \textbf{24.01\%}  & 0.676           & 0.557         & 0.622        & \textbf{11.68\%}  \\
\textbf{}          & \textbf{MRR}    & 0.631           & 0.378        & 0.556        & \textbf{47.33\%}  & 0.680           & 0.509         & 0.592        & \textbf{16.14\%}  \\\hline
\textbf{MPNN}      & \textbf{NDCG}   & 0.602           & 0.443        & 0.590        & \textbf{33.11\%}  & 0.646           & 0.307         & 0.621        & \textbf{101.92\%} \\
\textbf{}          & \textbf{MRR}    & 0.572           & 0.319        & 0.575        & \textbf{80.10\%}  & 0.660           & 0.145         & 0.595        & \textbf{311.42\%} \\\hline
\textbf{HMPNN}     & \textbf{NDCG}   & 0.789           & 0.399        & 0.623        & \textbf{56.14\%}  & 0.671           & 0.401         & 0.733        & \textbf{82.88\%}  \\
\textbf{}          & \textbf{MRR}    & 0.777           & 0.297        & 0.629        & \textbf{111.86\%} & 0.661           & 0.318         & 0.711        & \textbf{123.30\%} \\\hline\bottomrule
\end{tabular}
\normalsize
\vspace{3mm}
\end{table*}

\subsection{Generality of \method}
\label{sec:experiments:hgnn-types}

Here, we use $6$ different HGNN models, R-GCN~\cite{schlichtkrull2018modeling}, HAN~\cite{wang2019heterogeneous}, HGT~\cite{hu2020heterogeneous}, MAGNN~\cite{fu2020magnn}, MPNN~\cite{gilmer2017neural}, and HMPNN.
MPNN maps all node types to the shared embedding space using projection matrices at the beginning then applies MPNN layers designed for homogeneous graphs.
In Table~\ref{tab:hgnn}, \method improves accuracy on the target nodes across all HGNN models by up to $960\%$.
This shows the strong generality of \method.
More results and analysis can be found in Appendix~\ref{appendix:result:hgnn}.

\subsection{Sensitivity analysis}
\label{sec:experiments:sensitivity}

Using our synthetic heterogeneous graph generator, we generate non-trivial $2$-type heterogeneous graphs to examine how the feature and edge distributions affect the performance of \method and other baselines.
We generate a \emph{range} of test-case scenarios by manipulating (1) signal-to-noise ratio $\snre$ of within-class edge distributions and (2) signal-to-noise ratio $\snrf$ of within-class feature distributions across all of the (a) source-source ($s\leftrightarrow s$), (b) target-target ($t\leftrightarrow t$), and (c) source-target ($s\leftrightarrow t$) relationships. 

\begin{wrapfigure}{r}{0.5\textwidth}
 	\vspace{-3mm}
 	\includegraphics[width=0.5\textwidth]{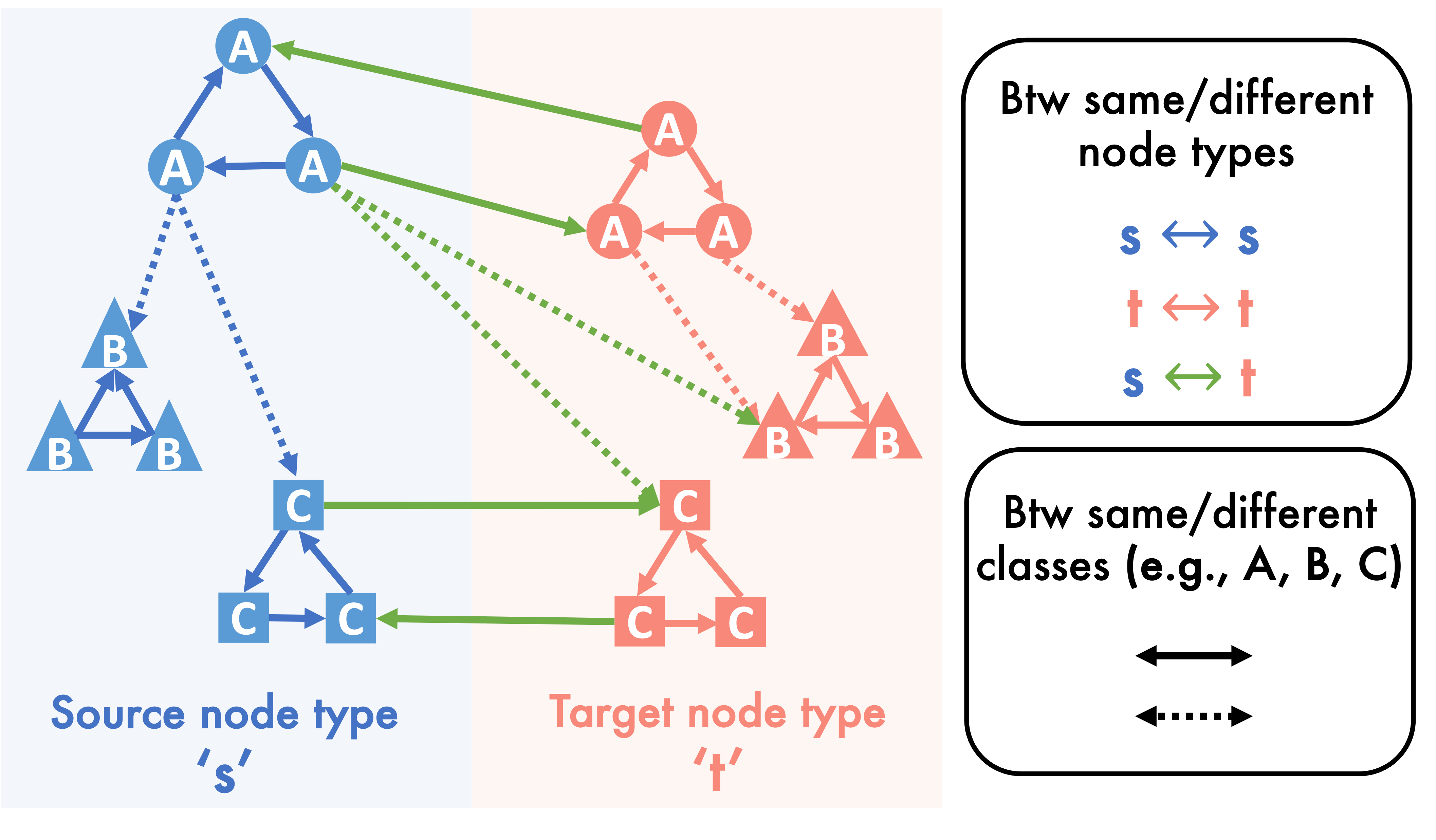}
 	\caption
 	{
 	    \small
 	    Synthetic HG generator
 	}
 	\label{fig:synthetic-generator}
 	\vspace{-3mm}
\end{wrapfigure}
For instance, in Figure~\ref{fig:synthetic-generator}, for each edge type ($s\leftrightarrow s$, $t\leftrightarrow t$, and $s\leftrightarrow t$, differentiated by colors), there are two different types of edges, edges within the same class (plain line) and edges across different classes (dotted line).
For each edge type, we manipulate $\snre$ by changing the ratio of within-class and cross-class edges, and $\snrf$ by diverging feature distributions between classes.
Thus there will be $6$ signal-to-noise ratios in total.
A higher signal-to-noise ratio for a particular data dimension (edges or features) across a particular relationship $r\in \{s\leftrightarrow s,\ t\leftrightarrow t,\ s\leftrightarrow t\}$ means that classes are more \emph{separable} in that data dimension, when comparing within $r$, and hence easier for HGNNs. 
Note that while tuning one $\sigma_{(\cdot)}$ on the range $[1.0, 10.0]$, the remaining five $\sigma_{(\cdot)}$ are held at $10.0$. Additionally, we vary $\sigma_{(\cdot)}$ across two scenarios: (I) ``easy": source and target node types have same number of clusters and same feature dimensions, (II) ``hard" source and target node types have different number of clusters and feature dimensions. 
Note that clusters and classes are different concepts in this experiment; several clusters could have the same class label.

\begin{figure}
 	\centering
 	\includegraphics[width=0.4\linewidth]{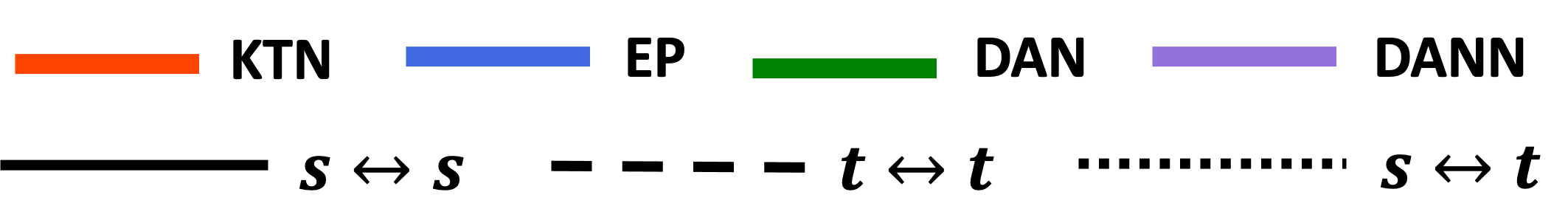} \\
 	\subfigure[Edge dist. (easy)]
 	{
 	\label{fig:2-type-simple:edge}
 	\includegraphics[width=0.23\linewidth]{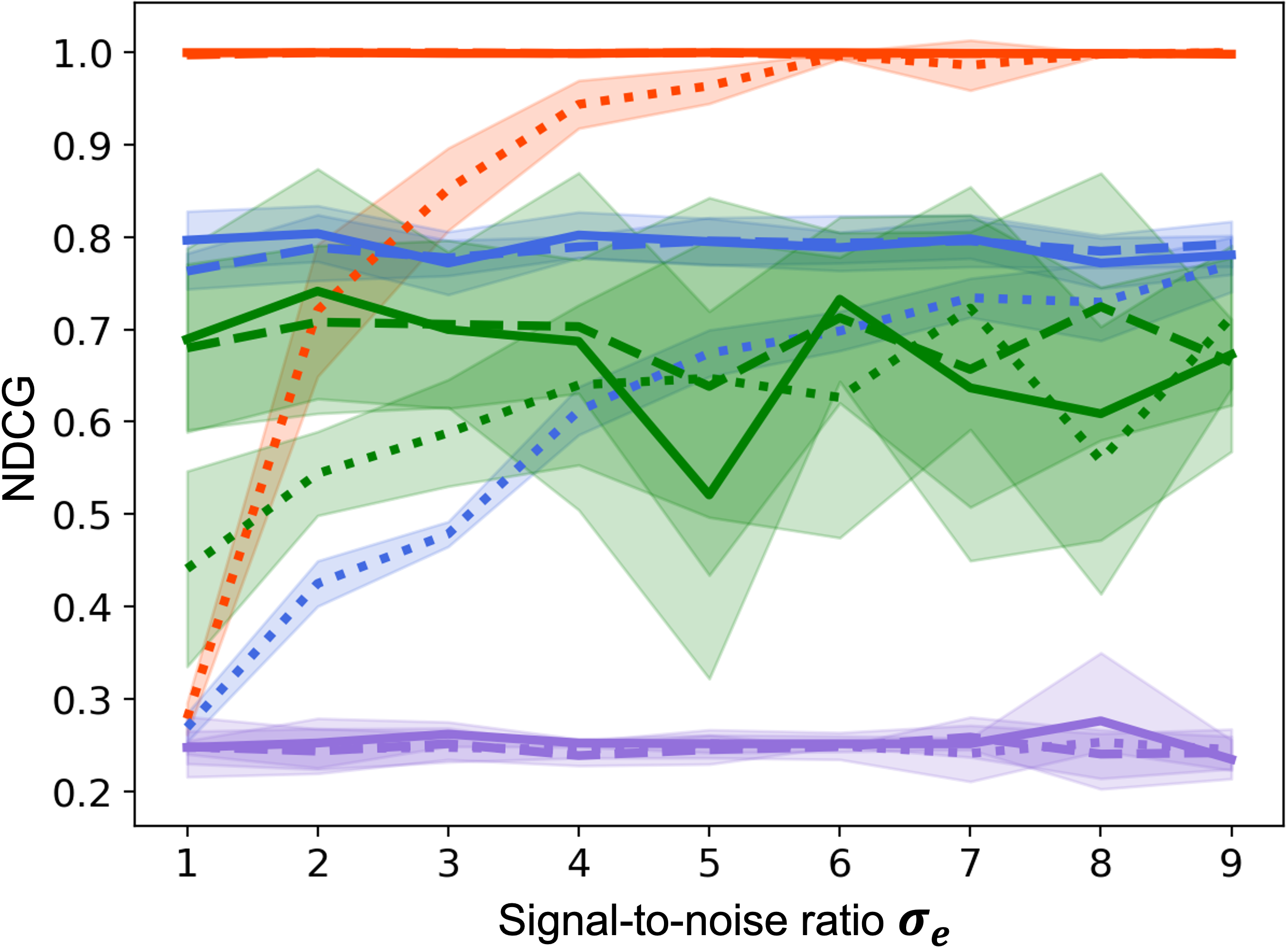}
 	}
 	\subfigure[Feature dist. (easy)]
 	{
 	\label{fig:2-type-simple:feat}
 	\includegraphics[width=.23\linewidth]{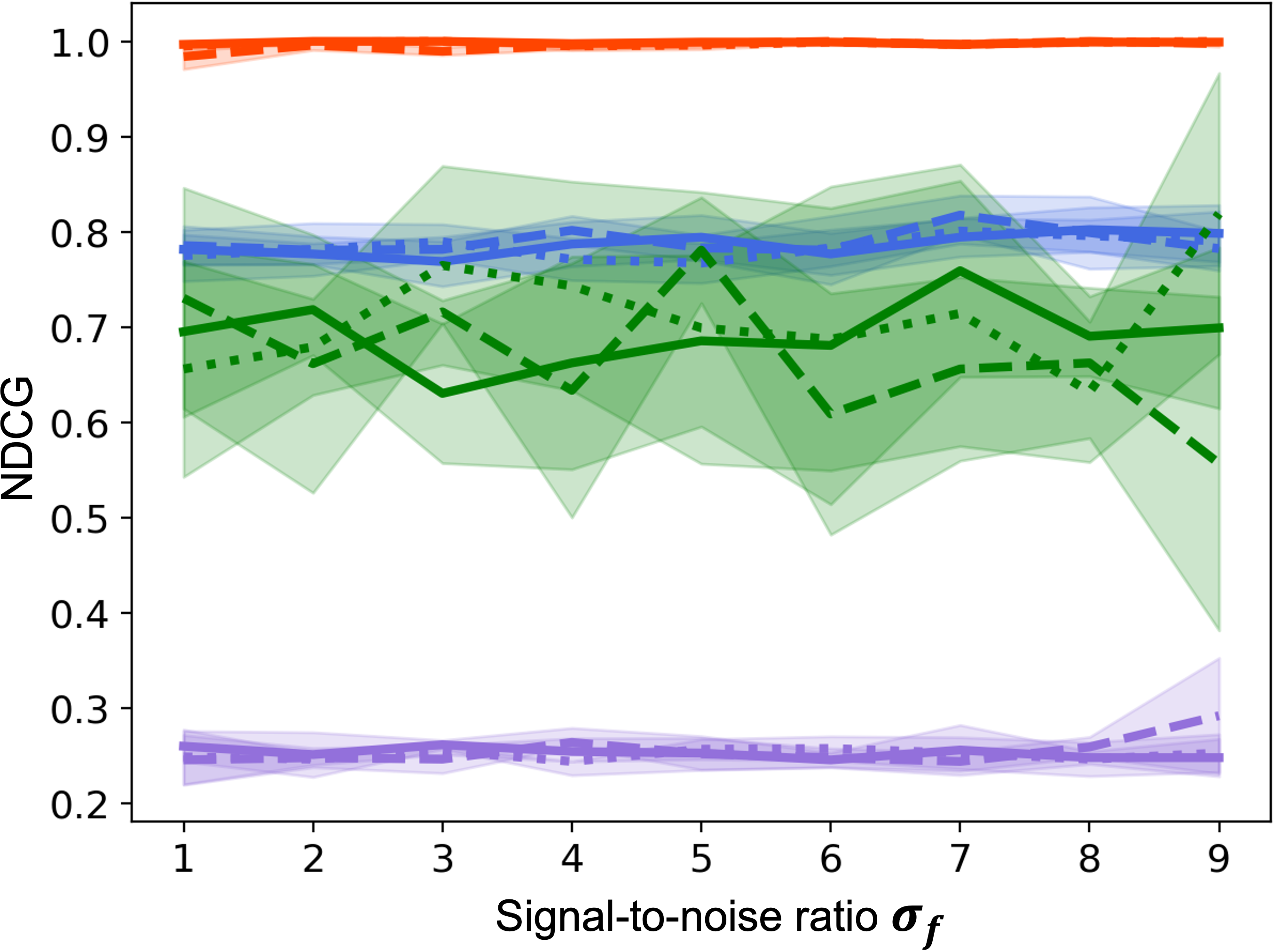}
 	}
 	\subfigure[Edge dist. (hard)]
 	{
 	\label{fig:2-type-hard:edge}
 	\includegraphics[width=0.23\linewidth]{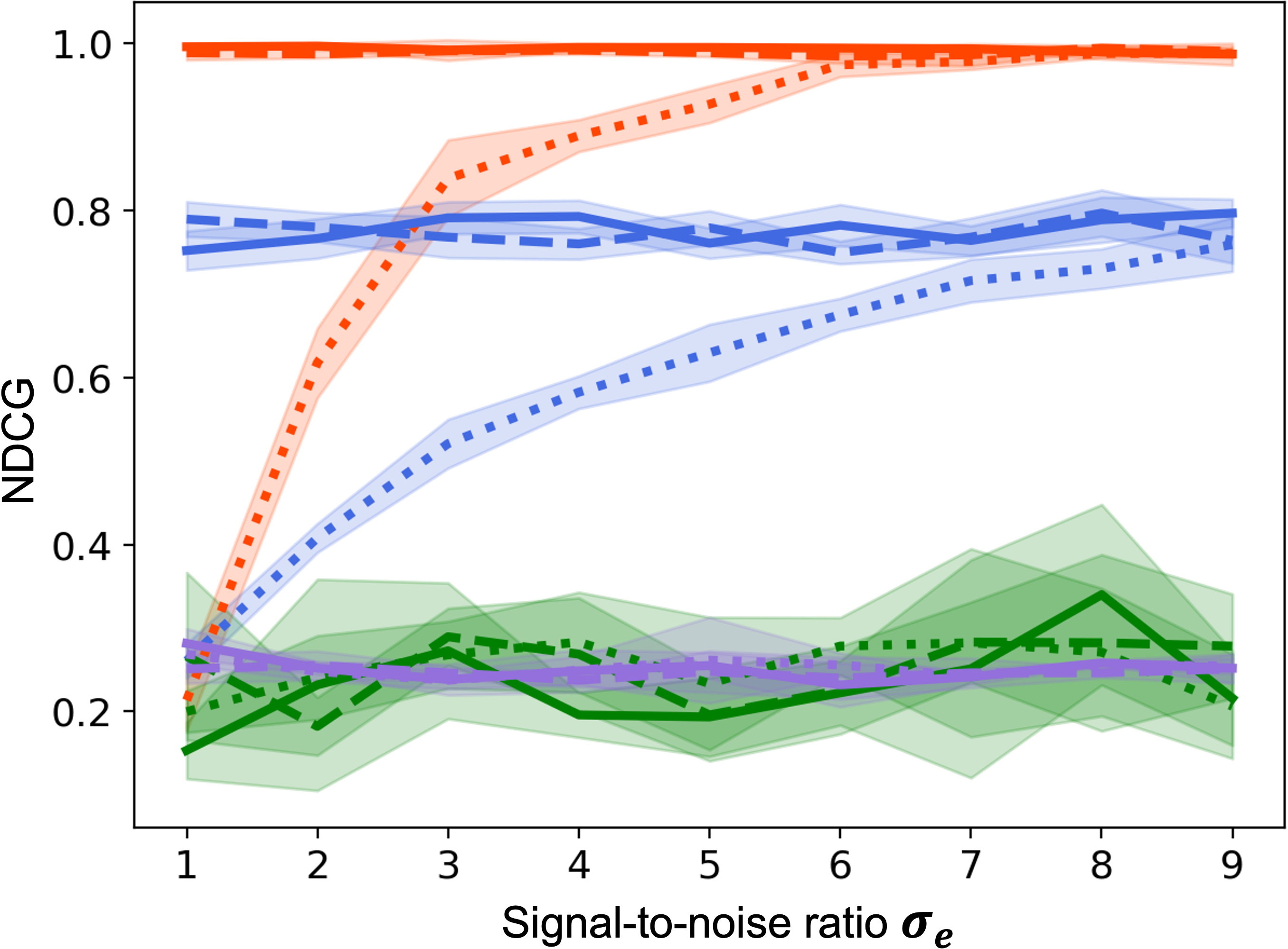}
 	}
 	\subfigure[Feature dist. (hard)]
 	{
 	\label{fig:2-type-hard:feat}
 	\includegraphics[width=.23\linewidth]{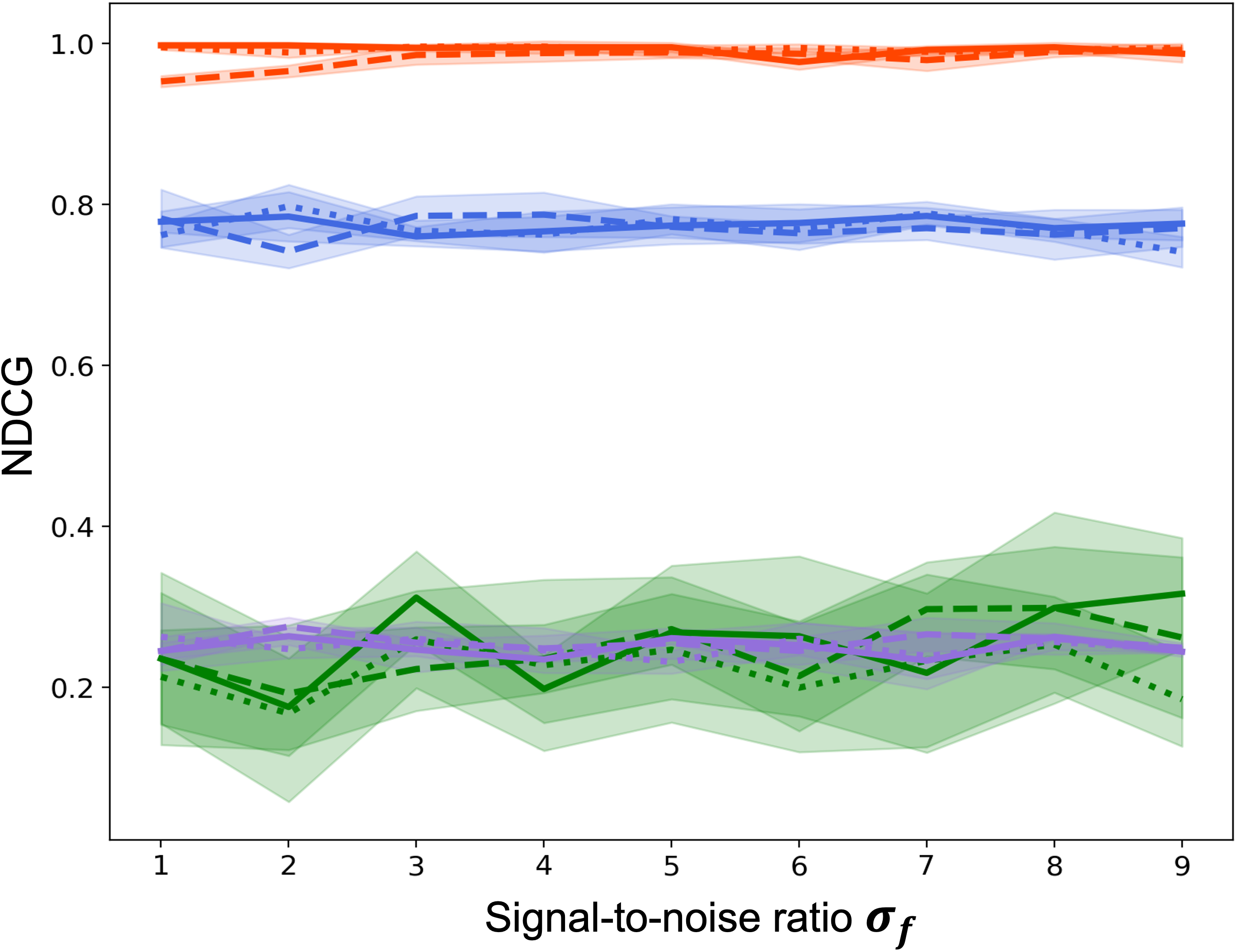}
 	}
 	\caption
 	{
 	    \small
 	    Effects of edge and feature distributions across classes and types in heterogeneous graphs.
 	}
 	\label{fig:2-type}
 \end{figure}

Figures~\ref{fig:2-type-simple:edge} and \ref{fig:2-type-hard:edge} show results from changing $\snre$ across the three relation types. We see that \method is affected only by $\snre$ across the $s\leftrightarrow t$ (cross-types) relationship, which accords with our theory, since \method exploits the between-type adjacency matrix. Surprisingly, as seen in Figures~\ref{fig:2-type-simple:feat} and \ref{fig:2-type-hard:feat}, we do not find a similar dependence of \method on $\snrf$, which shows that \method is robust by learning purely from edge homophily in the absence of feature homophily. 
Regarding the performance of other baselines, EP shows similar tendencies as \method --- only affected by cross-type $\snre$ --- because EP also relies on cross-type propagation along edges. However, its accuracy is bounded above due to the fact that it does not exploit the (unlabelled) target features.
DAN and DANN, which do not exploit cross-type edges, are not affected by cross-type $\snre$.
However, they show either low or unstable performance across different scenarios.
DAN shows especially poor performance in the ``hard" scenarios (Figure~\ref{fig:2-type-hard:edge} and~\ref{fig:2-type-hard:feat}), failing to deal with different feature spaces for source and target domains.
 
\section{Conclusion}
\label{sec:conclusion}
In this work, we proposed the first cross-type zero-shot transfer learning method for heterogeneous graphs.
Our method, Knowledge Transfer Networks (\method) for Heterogeneous Graph Neural Networks, transfers knowledge from \emph{label-abundant} node types to \emph{label-scarce} node types.
We illustrate \method handily improves HGNN performances up to $960\%$ for zero-labeled node types across $6$ different HGNN models and outperforms many challenging baselines up to $73\%$ higher in MRR.
Future work in the area includes filtering noisy edges between source and target domains and making \method more robust and less dependent on structure of given noisy heterogeneous graphs.
\vspace{-5pt}
\paragraph{Limitation Statement} 
Our transfer learning method is limited to node types sharing the same task (i.e., the same classifier).
We plan to extend our work to transfer knowledge between different tasks running on different node types on a heterogeneous graph. 
\vspace{-5pt}
\paragraph{Societal Impact Statement}  
\method allows organizations to learn better from their own graph data, leveraging its structure without requiring external information.
We believe this has a number of positive applications (preserving model quality without needing extra datasets).
However like all modeling improvements, its true impact depends on what modeling tasks the technique is applied to.

\section{Acknowledgement}
\label{sec:acknowledgement}
MY gratefully acknowledges support from Amazon Graduate Research Fellowship.
GPUs are partially supported by AWS Cloud Credit for Research program.

\bibliography{myref}

\begin{thebibliography}{10}

\bibitem{abbe2017community}
Emmanuel Abbe.
\newblock Community detection and stochastic block models: recent developments.
\newblock {\em The Journal of Machine Learning Research}, 18(1):6446--6531,
  2017.

\bibitem{ben2010theory}
Shai Ben-David, John Blitzer, Koby Crammer, Alex Kulesza, Fernando Pereira, and
  Jennifer~Wortman Vaughan.
\newblock A theory of learning from different domains.
\newblock {\em Machine learning}, 79(1):151--175, 2010.

\bibitem{ben2007analysis}
Shai Ben-David, John Blitzer, Koby Crammer, Fernando Pereira, et~al.
\newblock Analysis of representations for domain adaptation.
\newblock {\em Advances in neural information processing systems}, 19:137,
  2007.

\bibitem{bordes2013translating}
Antoine Bordes, Nicolas Usunier, Alberto Garcia-Duran, Jason Weston, and Oksana
  Yakhnenko.
\newblock Translating embeddings for modeling multi-relational data.
\newblock {\em Advances in neural information processing systems}, 26, 2013.

\bibitem{dong2017metapath2vec}
Yuxiao Dong, Nitesh~V Chawla, and Ananthram Swami.
\newblock metapath2vec: Scalable representation learning for heterogeneous
  networks.
\newblock In {\em Proceedings of the 23rd ACM SIGKDD international conference
  on knowledge discovery and data mining}, pages 135--144, 2017.

\bibitem{ferludin2022tfgnn}
Oleksandr Ferludin, Arno Eigenwillig, Martin Blais, Dustin Zelle, Jan Pfeifer,
  Alvaro Sanchez-Gonzalez, Sibon Li, Sami Abu-El-Haija, Peter Battaglia,
  Neslihan Bulut, et~al.
\newblock Tf-gnn: Graph neural networks in tensorflow.
\newblock {\em arXiv preprint arXiv:2207.03522}, 2022.

\bibitem{fey2019fast}
Matthias Fey and Jan~Eric Lenssen.
\newblock Fast graph representation learning with pytorch geometric.
\newblock {\em arXiv preprint arXiv:1903.02428}, 2019.

\bibitem{fu2020magnn}
Xinyu Fu, Jiani Zhang, Ziqiao Meng, and Irwin King.
\newblock Magnn: Metapath aggregated graph neural network for heterogeneous
  graph embedding.
\newblock In {\em Proceedings of The Web Conference 2020}, pages 2331--2341,
  2020.

\bibitem{ganin2016domain}
Yaroslav Ganin, Evgeniya Ustinova, Hana Ajakan, Pascal Germain, Hugo
  Larochelle, Fran{\c{c}}ois Laviolette, Mario Marchand, and Victor Lempitsky.
\newblock Domain-adversarial training of neural networks.
\newblock {\em The journal of machine learning research}, 17(1):2096--2030,
  2016.

\bibitem{gilmer2017neural}
Justin Gilmer, Samuel~S Schoenholz, Patrick~F Riley, Oriol Vinyals, and
  George~E Dahl.
\newblock Neural message passing for quantum chemistry.
\newblock In {\em International conference on machine learning}, pages
  1263--1272. PMLR, 2017.

\bibitem{gretton2012kernel}
Arthur Gretton, Karsten~M Borgwardt, Malte~J Rasch, Bernhard Sch{\"o}lkopf, and
  Alexander Smola.
\newblock A kernel two-sample test.
\newblock {\em The Journal of Machine Learning Research}, 13(1):723--773, 2012.

\bibitem{hamilton2017inductive}
William~L Hamilton, Rex Ying, and Jure Leskovec.
\newblock Inductive representation learning on large graphs.
\newblock In {\em Proceedings of the 31st International Conference on Neural
  Information Processing Systems}, pages 1025--1035, 2017.

\bibitem{hu2019strategies}
Weihua Hu, Bowen Liu, Joseph Gomes, Marinka Zitnik, Percy Liang, Vijay Pande,
  and Jure Leskovec.
\newblock Strategies for pre-training graph neural networks.
\newblock {\em arXiv preprint arXiv:1905.12265}, 2019.

\bibitem{hu2020gpt}
Ziniu Hu, Yuxiao Dong, Kuansan Wang, Kai-Wei Chang, and Yizhou Sun.
\newblock Gpt-gnn: Generative pre-training of graph neural networks.
\newblock In {\em Proceedings of the 26th ACM SIGKDD International Conference
  on Knowledge Discovery \& Data Mining}, pages 1857--1867, 2020.

\bibitem{hu2020heterogeneous}
Ziniu Hu, Yuxiao Dong, Kuansan Wang, and Yizhou Sun.
\newblock Heterogeneous graph transformer.
\newblock In {\em Proceedings of The Web Conference 2020}, pages 2704--2710,
  2020.

\bibitem{huang2020hgt}
Tiancheng Huang, Ke~Xu, and Donglin Wang.
\newblock Da-hgt: Domain adaptive heterogeneous graph transformer.
\newblock {\em arXiv preprint arXiv:2012.05688}, 2020.

\bibitem{kipf2016semi}
Thomas~N Kipf and Max Welling.
\newblock Semi-supervised classification with graph convolutional networks.
\newblock {\em arXiv preprint arXiv:1609.02907}, 2016.

\bibitem{long2015learning}
Mingsheng Long, Yue Cao, Jianmin Wang, and Michael Jordan.
\newblock Learning transferable features with deep adaptation networks.
\newblock In {\em International conference on machine learning}, pages 97--105.
  PMLR, 2015.

\bibitem{long2017conditional}
Mingsheng Long, Zhangjie Cao, Jianmin Wang, and Michael~I Jordan.
\newblock Conditional adversarial domain adaptation.
\newblock {\em arXiv preprint arXiv:1705.10667}, 2017.

\bibitem{long2017deep}
Mingsheng Long, Han Zhu, Jianmin Wang, and Michael~I Jordan.
\newblock Deep transfer learning with joint adaptation networks.
\newblock In {\em International conference on machine learning}, pages
  2208--2217. PMLR, 2017.

\bibitem{luo2020progressive}
Yadan Luo, Zijian Wang, Zi~Huang, and Mahsa Baktashmotlagh.
\newblock Progressive graph learning for open-set domain adaptation.
\newblock In {\em International Conference on Machine Learning}, pages
  6468--6478. PMLR, 2020.

\bibitem{ma2019gcan}
Xinhong Ma, Tianzhu Zhang, and Changsheng Xu.
\newblock Gcan: Graph convolutional adversarial network for unsupervised domain
  adaptation.
\newblock In {\em Proceedings of the IEEE/CVF Conference on Computer Vision and
  Pattern Recognition}, pages 8266--8276, 2019.

\bibitem{mikolov2013distributed}
Tomas Mikolov, Ilya Sutskever, Kai Chen, Greg~S Corrado, and Jeff Dean.
\newblock Distributed representations of words and phrases and their
  compositionality.
\newblock In {\em Advances in neural information processing systems}, pages
  3111--3119, 2013.

\bibitem{qiu2020gcc}
Jiezhong Qiu, Qibin Chen, Yuxiao Dong, Jing Zhang, Hongxia Yang, Ming Ding,
  Kuansan Wang, and Jie Tang.
\newblock Gcc: Graph contrastive coding for graph neural network pre-training.
\newblock In {\em Proceedings of the 26th ACM SIGKDD International Conference
  on Knowledge Discovery \& Data Mining}, pages 1150--1160, 2020.

\bibitem{redko2017theoretical}
Ievgen Redko, Amaury Habrard, and Marc Sebban.
\newblock Theoretical analysis of domain adaptation with optimal transport.
\newblock In {\em Joint European Conference on Machine Learning and Knowledge
  Discovery in Databases}, pages 737--753. Springer, 2017.

\bibitem{schlichtkrull2018modeling}
Michael Schlichtkrull, Thomas~N Kipf, Peter Bloem, Rianne Van Den~Berg, Ivan
  Titov, and Max Welling.
\newblock Modeling relational data with graph convolutional networks.
\newblock In {\em European semantic web conference}, pages 593--607. Springer,
  2018.

\bibitem{shen2018wasserstein}
Jian Shen, Yanru Qu, Weinan Zhang, and Yong Yu.
\newblock Wasserstein distance guided representation learning for domain
  adaptation.
\newblock In {\em Thirty-Second AAAI Conference on Artificial Intelligence},
  2018.

\bibitem{sinha2015overview}
Arnab Sinha, Zhihong Shen, Yang Song, Hao Ma, Darrin Eide, Bo-June Hsu, and
  Kuansan Wang.
\newblock An overview of microsoft academic service (mas) and applications.
\newblock In {\em Proceedings of the 24th international conference on world
  wide web}, pages 243--246, 2015.

\bibitem{sun2016return}
Baochen Sun, Jiashi Feng, and Kate Saenko.
\newblock Return of frustratingly easy domain adaptation.
\newblock In {\em Proceedings of the AAAI Conference on Artificial
  Intelligence}, volume~30, 2016.

\bibitem{sun2012mining}
Yizhou Sun and Jiawei Han.
\newblock Mining heterogeneous information networks: principles and
  methodologies.
\newblock {\em Synthesis Lectures on Data Mining and Knowledge Discovery},
  3(2):1--159, 2012.

\bibitem{tang2008arnetminer}
Jie Tang, Jing Zhang, Limin Yao, Juanzi Li, Li~Zhang, and Zhong Su.
\newblock Arnetminer: extraction and mining of academic social networks.
\newblock In {\em Proceedings of the 14th ACM SIGKDD international conference
  on Knowledge discovery and data mining}, pages 990--998, 2008.

\bibitem{tsitsulin2020graph}
Anton Tsitsulin, John Palowitch, Bryan Perozzi, and Emmanuel M{\"u}ller.
\newblock Graph clustering with graph neural networks.
\newblock {\em arXiv preprint arXiv:2006.16904}, 2020.

\bibitem{tsitsulin2021synthetic}
Anton Tsitsulin, Benedek Rozemberczki, John Palowitch, and Bryan Perozzi.
\newblock Synthetic graph generation to benchmark graph learning.
\newblock {\em WWW'21, Workshop on Graph Learning Benchmarks}, 2021.

\bibitem{wang2019deep}
Minjie Wang, Da~Zheng, Zihao Ye, Quan Gan, Mufei Li, Xiang Song, Jinjing Zhou,
  Chao Ma, Lingfan Yu, Yu~Gai, et~al.
\newblock Deep graph library: A graph-centric, highly-performant package for
  graph neural networks.
\newblock {\em arXiv preprint arXiv:1909.01315}, 2019.

\bibitem{wang2019heterogeneous}
Xiao Wang, Houye Ji, Chuan Shi, Bai Wang, Yanfang Ye, Peng Cui, and Philip~S
  Yu.
\newblock Heterogeneous graph attention network.
\newblock In {\em The World Wide Web Conference}, pages 2022--2032, 2019.

\bibitem{wolf2020transformers}
Thomas Wolf, Julien Chaumond, Lysandre Debut, Victor Sanh, Clement Delangue,
  Anthony Moi, Pierric Cistac, Morgan Funtowicz, Joe Davison, Sam Shleifer,
  et~al.
\newblock Transformers: State-of-the-art natural language processing.
\newblock In {\em Proceedings of the 2020 Conference on Empirical Methods in
  Natural Language Processing: System Demonstrations}, pages 38--45, 2020.

\bibitem{wu2020unsupervised}
Man Wu, Shirui Pan, Chuan Zhou, Xiaojun Chang, and Xingquan Zhu.
\newblock Unsupervised domain adaptive graph convolutional networks.
\newblock In {\em Proceedings of The Web Conference 2020}, pages 1457--1467,
  2020.

\bibitem{wu2021towards}
Qitian Wu, Chenxiao Yang, and Junchi Yan.
\newblock Towards open-world feature extrapolation: An inductive graph learning
  approach.
\newblock {\em Advances in Neural Information Processing Systems},
  34:19435--19447, 2021.

\bibitem{yang2020heterogeneous}
Carl Yang, Yuxin Xiao, Yu~Zhang, Yizhou Sun, and Jiawei Han.
\newblock Heterogeneous network representation learning: A unified framework
  with survey and benchmark.
\newblock {\em IEEE Transactions on Knowledge and Data Engineering}, 2020.

\bibitem{yang2021domain}
Shuwen Yang, Guojie Song, Yilun Jin, and Lun Du.
\newblock Domain adaptive classification on heterogeneous information networks.
\newblock In {\em Proceedings of the Twenty-Ninth International Conference on
  International Joint Conferences on Artificial Intelligence}, pages
  1410--1416, 2021.

\bibitem{yang2021interpretable}
Yaming Yang, Ziyu Guan, Jianxin Li, Wei Zhao, Jiangtao Cui, and Quan Wang.
\newblock Interpretable and efficient heterogeneous graph convolutional
  network.
\newblock {\em IEEE Transactions on Knowledge and Data Engineering}, 2021.

\bibitem{you2020handling}
Jiaxuan You, Xiaobai Ma, Yi~Ding, Mykel~J Kochenderfer, and Jure Leskovec.
\newblock Handling missing data with graph representation learning.
\newblock {\em Advances in Neural Information Processing Systems},
  33:19075--19087, 2020.

\bibitem{zhang2019heterogeneous}
Chuxu Zhang, Dongjin Song, Chao Huang, Ananthram Swami, and Nitesh~V Chawla.
\newblock Heterogeneous graph neural network.
\newblock In {\em Proceedings of the 25th ACM SIGKDD International Conference
  on Knowledge Discovery \& Data Mining}, pages 793--803, 2019.

\bibitem{zhang2019oag}
Fanjin Zhang, Xiao Liu, Jie Tang, Yuxiao Dong, Peiran Yao, Jie Zhang, Xiaotao
  Gu, Yan Wang, Bin Shao, Rui Li, et~al.
\newblock Oag: Toward linking large-scale heterogeneous entity graphs.
\newblock In {\em Proceedings of the 25th ACM SIGKDD International Conference
  on Knowledge Discovery \& Data Mining}, pages 2585--2595, 2019.

\bibitem{zhu2005semi}
Xiaojin Zhu.
\newblock {\em Semi-supervised learning with graphs}.
\newblock Carnegie Mellon University, 2005.

\end{thebibliography}
\bibliographystyle{plain}

\section*{Checklist}
\begin{enumerate}

\item For all authors...
\begin{enumerate}
    \item Do the main claims made in the abstract and introduction accurately reflect the paper's contributions and scope?
    \answerYes{}
    \item Did you describe the limitations of your work?
    \answerYes{We describe it in Section~\ref{sec:conclusion}. We also show which environments are ideal for our method via sensitivity test in Section~\ref{sec:experiments:sensitivity}.}
    \item Did you discuss any potential negative societal impacts of your work?
    \answerYes{We describe it in Section~\ref{sec:conclusion}. We also mention about possible private information leakage in graph-to-graph transfer learning in Section~\ref{sec:introduction}. By proposing transfer learning between node types in one heterogeneous graph, we can alleviate this issue.}
    \item Have you read the ethics review guidelines and ensured that your paper conforms to them?
    \answerYes{}
\end{enumerate}

\item If you are including theoretical results...
\begin{enumerate}
    \item Did you state the full set of assumptions of all theoretical results?
    \answerYes{We mention the full set of assumptions for Theorem~\ref{theorem} in Section~\ref{sec:motivation:theoretical_analysis}.}
    \item Did you include complete proofs of all theoretical results?
    \answerYes{The proof for Theorem~\ref{theorem} is described in Appendix~\ref{appendix:theorem1}.}
\end{enumerate}

\item If you ran experiments...
\begin{enumerate}
    \item Did you include the code, data, and instructions needed to reproduce the main experimental results (either in the supplemental material or as a URL)?
    \answerYes{We provide a URL to our code in Appendix~\ref{appendix:experiment-setting}.}
    \item Did you specify all the training details (e.g., data splits, hyperparameters, how they were chosen)?
    \answerYes{They are specified in Appendix~\ref{appendix:dataset} and~\ref{appendix:experiment-setting}.}
    \item Did you report error bars (e.g., with respect to the random seed after running experiments multiple times)?
    \answerYes{They are specified in Appendix~\ref{appendix:result:da} and~\ref{appendix:result:hgnn}.}
    \item Did you include the total amount of compute and the type of resources used (e.g., type of GPUs, internal cluster, or cloud provider)?
    \answerYes{They are specified in Appendix~\ref{appendix:experiment-setting}.}
\end{enumerate}

\item If you are using existing assets (e.g., code, data, models) or curating/releasing new assets...
\begin{enumerate}
    \item If your work uses existing assets, did you cite the creators?
    \answerYes{We cite public domain adaptation library ADA and heterogeneous graph neural network library OpenHGNN that we have used in our experiment in Appendix~\ref{appendix:experiment-setting}.}
    \item Did you mention the license of the assets?
    \answerNA{All codes and data we used are public.}
    \item Did you include any new assets either in the supplemental material or as a URL?
    \answerYes{We provide a URL to our code in Appendix~\ref{appendix:experiment-setting}.}
    \item Did you discuss whether and how consent was obtained from people whose data you're using/curating?
    \answerNA{We do not use any personal data.}
    \item Did you discuss whether the data you are using/curating contains personally identifiable information or offensive content?
    \answerNA{We do not use any personal data.}
\end{enumerate}

\item If you used crowdsourcing or conducted research with human subjects...
\begin{enumerate}
    \item Did you include the full text of instructions given to participants and screenshots, if applicable?
    \answerNA{}
    \item Did you describe any potential participant risks, with links to Institutional Review Board (IRB) approvals, if applicable?
    \answerNA{}
    \item Did you include the estimated hourly wage paid to participants and the total amount spent on participant compensation?
    \answerNA{}
\end{enumerate}

\end{enumerate}


\newpage

\appendix
\section{Appendix}
\label{sec:appendix}
\subsection{Proof of Theorem~\ref{theorem}}\label{appendix:theorem1}
In this proof, we adopt a simplified version of our message-passing function that ignores the skip-connection:
\begin{equation}
    \small
    \textbf{Message}^{(l)}(i, j) = M_{\phi(i,j)}^{(l)}h_i^{(j)}.
\end{equation}
The HGNN trained in the experimental results shown in Figure~\ref{fig:toy_exp} also does not use skip-connections and hence represents a theoretically-exact KTN component. In the real experiments, we use (1) skip-connections, exploiting their usual benefits~\cite{hamilton2017inductive}, and (2) the trainable version of KTN.

\begin{proof}
Without loss of generality, we prove the result for the case where $\mathcal{R} = \{(s, t): s,t\in\mathcal{T}\}$, meaning the type of an edge is identified with the (ordered) types of the neighbor nodes. In other words, there is only one edge modality possible, such as a social networks with multiple node types (e.g.\ ``users", ``groups") but only one edge modality (``friendship"). In the case of multiple edge modalities (e.g. ``friendship" and ``message"), the result is extended trivially (through with more algebraically-dense forms of $a_{ts}$ and $q_{ts}$).

Throughout this proof, we use the following notation for the set of all $j$-adjacent edges of relation type $r$:
\begin{equation}
    \small
    \mathcal{E}_r(j):=\{(i,j): i\in\mathcal{V}, (i,j) = r\}.
\end{equation}
We write $A_{x_1x_2}$ to denote the sub-matrix of the total $n_{x_1}\times n_{x_2}$ adjacency matrix $A$ corresponding to node types $x_1,x_2\in\mathcal{T}$, and $\bar{A}_{x_1x_2}$ to denote the same matrix divided by its column sum. $H_x^{(l)}$ is the (row-wise) $n_x\times d_l$ embedding matrix of $x$-type nodes at layer $l$.

We first compute the $l$-\emph{th} output $g_j^{(l)}$ of the $\textbf{Aggregate}$ step defined for HGNNs in Equation~\ref{eq:aggregate}, for any node $j\in\mathcal{V}$ such that $\tau(j) = s$. The output of \textbf{Aggregate} is a concatenation of edge-type-specific aggregations (see Equation~\ref{eq:aggregate}). Note that at most $T = |\mathcal{T}|$ elements of this concatenation are non-zero, since the node $j$ only participates in $T$ out of $T^2$ relation types in $\mathcal{R}$. Thus we can write $g_j^{(l)}$ as 
\begin{align*}
    \small
    g_j^{(l)} &= \underset{r\in\mathcal{R}}{\concat}\tfrac{1}{|\mathcal{E}_r(j)|}\sum_{e\in\mathcal{E}_r(j)}\textbf{Message}^{(l)}(e)\\
    &= \underset{x\in\mathcal{T}}{\concat}\tfrac{1}{|\mathcal{E}_{xs}(j)|}\sum_{e\in\mathcal{E}_{xs}(j)}\textbf{Message}^{(l)}(e)\\
    &=\underset{x\in\mathcal{T}}{\concat}\tfrac{1}{|\mathcal{E}_{xs}(j)|}\sum_{(i,j)\in\mathcal{E}_{xs}(j)}M_{xs}^{(l)}h_i^{(l-1)}\\
    &=\underset{x\in\mathcal{T}}{\concat}\tfrac{1}{|\mathcal{E}_{xs}(j)|}M_{xs}^{(l)}\sum_{(i,j)\in\mathcal{E}_{xs}(j)}h_i^{(l-1)}\\
    &=\underset{x\in\mathcal{T}}{\concat}M_{xs}^{(l)}\left(H_x^{(l-1)}\right)'\bar{A}_{xs}^{(j)},
\end{align*}
where $\bar{A}_{xs}^{(j)}$ denotes the $j$-\emph{th} column of $\bar{A}_{xs}$. Notice that
\begin{equation}
    \small
    h_j^{(l)} = \textbf{Transform}^{(l)}(j) = W_s^{(l)}g_j^{(l)},
\end{equation}
and (again) at most $T$ elements of the concatenation $g_j^{(l)}$ are non-zero. Therefore let $W_{xs}^{(l)}$ be the columns of $W_s^{(l)}$ that select the concatenated element of $g_j^{(l)}$ corresponding to node type $x$. Then we can write
\begin{equation}
    \small
    h_j^{(l)} = \sum_{x\in\mathcal{T}}W_{xs}^{(l)}M_{xs}^{(l)}\left(H_x^{(l-1)}\right)'\bar{A}_{xs}^{(j)}.
\end{equation}

\begin{algorithm}[t!]
    \caption{Training step for one minibatch (indirect version)}
    \label{alg:train-extend}
\begin{algorithmic}[1]
\small
    \REQUIRE heterogeneous graph $\mathcal{G} = (\mathcal{V}, \mathcal{E}, \mathcal{T}, \mathcal{R})$, node feature matrices $X$, adjacency matrices $A_{xy}$ where $\forall(x , y)\in \mathcal{R}$, source node type $s$, target node type $t$, source node label matrix $Y_s$.
    \ENSURE HGNN $\textbf{f}$, classifier $\textbf{g}$, \method $\textbf{t}_{\text{KTN}}$
    \STATE $H^{(L)}_s, H^{(L)}_t = \textbf{f}(H^{(0)} = X, \mathcal{G})$, $H^{*}_{t} = \textbf{0}$ 
    \FOR{each meta-path $p = t \rightarrow s$} 
        \STATE $x = t$, $Z = H^{(L)}_t$
        \FOR{each node type $y \in p$}
            \STATE $Z = A_{xy}ZT_{xy}$
            \STATE $x = y$
        \ENDFOR 
        \STATE $H^{*}_{t} = H^{*}_{t} + Z$
    \ENDFOR 
    \STATE $\mathcal{L}_{\text{KTN}} = \left\|H^{(L)}_{s} - H^{*}_{t}\right\|_{2}$
    \STATE $\mathcal{L} = \mathcal{L}_{\text{CL}}(\textbf{g}(H^{(L)}_s), Y_s) + \lambda\mathcal{L}_{\text{KTN}}$
    \STATE Update $\textbf{f}$, $\textbf{g}$, $\textbf{t}_{\text{KTN}}$ using $\nabla\mathcal{L}$
\normalsize
\end{algorithmic}
\end{algorithm}

\begin{algorithm}[t!]
    \caption{Test step for a target domain (indirect version)}
    \label{alg:test-extend}
\begin{algorithmic}[1]
\small
    \REQUIRE pretrained HGNN $\textbf{f}$, classifier $\textbf{g}$, \method $\textbf{t}_{\text{KTN}}$
    \ENSURE target node label matrix $Y_t$
    \STATE $H^{(L)}_t = \textbf{f}(H^{(0)} = X, \mathcal{G})$, $H^{*}_{t} = \textbf{0}$
    \FOR{each meta-path $p = t \rightarrow s$}
    \STATE $x = t$, $Z = H^{(L)}_t$
    \FOR{each node type $y \in p$}
    \STATE $X = ZT_{xy}$
    \STATE $x = y$
    \ENDFOR
    \STATE $H^{*}_{t} = H^{*}_{t} + Z$
    \ENDFOR
    \RETURN $\textbf{g}(H^{*}_{t})$
\normalsize
\end{algorithmic}
\end{algorithm}

Defining the operator $Q_{xs}^{(l)} := \left(W_{xs}^{(l)}M_{xs}^{(l)}\right)'$, this implies that
\begin{align*}
    \small
    &H^{(l)}_s = \sum_{x\in\mathcal{T}}\bar{A}_{xs}H_x^{(l-1)}Q_{xs}^{(l-1)} \\
    &= [\bar{A}_{x_{1}s},\ldots,\bar{A}_{x_{T}s}]
    \begin{bmatrix}
        H^{(l-1)}_{x_1} & 0 & 0\\
        0 & \ldots & 0\\
        0 & 0 & H^{(l-1)}_{x_T}
    \end{bmatrix}
    \begin{bmatrix}
        Q_{x_1s}^{(l-1)}\\
        \ldots\\
        Q_{x_Ts}^{(l-1)}
   \end{bmatrix}
   \\
   & = \bar{A}_{\cdot s}H_{\cdot}^{(l-1)}Q_{\cdot s}^{(l-1)}
\end{align*}
Similarly we have $H^{(l)}_t = \bar{A}_{\cdot t}H_{\cdot}^{(l-1)}Q_{\cdot t}^{(l-1)}$. Since $H^{(l)}_s$ and $H^{(l)}_t$ share the term $H_\cdot^{(l-1)}$, we can write
\begin{equation}
\label{eq:thoretical}
    \small
    H_s^{(l)} = \bar{A}_{\cdot s}\bar{A}^{-1}_{\cdot t} H^{(l)}_{t} (Q_{\cdot t}^{(l-1)})^{-1}Q_{\cdot s}^{(l-1)},
\end{equation}
where $X^{-1}$ denotes the pseudo-inverse.
\end{proof}

\subsection{Indirectly Connected Source and Target Node Types}
\label{appendix:indirect}

When source and target node types are indirectly connected by another node type $x$, we can simply extend $\textbf{t}_{\text{KTN}}(H^{(L)}_{t})$ to $(A_{xs}(A_{tx}H^{(L)}_{t}T_{tx})T_{xs})$ where $T_{tx}T_{xs}$ becomes a mapping function from target to source domains.
Algorithms~\ref{alg:train-extend} and~\ref{alg:test-extend} show how to extend \method.
For every step ($x \rightarrow y$) in a meta-path ($t \rightarrow \cdots \rightarrow s$) connecting target node type $t$ to source node type $s$, we define a transformation matrix $T_{xy}$, run a convolution operation with an adjacency matrix $A_{xy}$, then map the transformed embedding to the source domain.
We run the same process for all meta-paths connecting from target node type $t$ to source node type $s$, and sum up them to match with the source embeddings.
In the test phase, we run the same process to get the transformed target embeddings, but this time, without adjacency matrices.
We run Algorithm~\ref{alg:train-extend} and~\ref{alg:test-extend} for transfer learning tasks between author and venue nodes which are indirectly connected by paper nodes in OAG graphs (Figure~\ref{fig:schema1:oag}).
As shown in Tables~\ref{tab:oag:cs2},~\ref{tab:oag:cn}, and~\ref{tab:oag:ml}, we successfully transfer HGNN models between author and venue nodes (A-V and V-A) for both L1 and L2 tasks.

Will lengths of meta-paths affect the performance?
We examine the performance of \method varying the length of meta-paths between source and target node types.
In Table~\ref{tab:meta-path}, accuracy decreases with longer meta-paths.
When we add additional meta-paths than the minimum path, it also brings noise in every edge types.
Note that author and venue nodes are indirectly connected by paper nodes; thus the minimum length of meta-paths in the A-V (L1) task is $2$.
The accuracy in the A-V (L1) task with a meta-path of length $1$ is low because \method fails to transfer anything with a meta-path shorter than the minimum.
Using the minimum length of meta-paths is enough for \method.

\begin{table*}[t]
    \caption{
    \small
	\textbf{Open Academic Graph on Computer Science field}.  
	The \textit{gain} column shows the relative gain of our method over using no domain adaptation (\textit{Base} column).
	\textit{o.o.m} denotes \textit{out-of-memory} errors.
	}
	\label{tab:oag:cs2}
	\centering
    \tiny
\begin{tabular}{l|l|c|cc|ccc|c|cc|l} \toprule\hline
\multicolumn{1}{c}{\textbf{Task}} & \multicolumn{1}{c}{\textbf{Metric}} & \textbf{Base} & \textbf{DAN} & \textbf{JAN} & \textbf{DANN} & \textbf{CDAN} & \textbf{CDAN-E} & \textbf{WDGRL} & \textbf{LP} & \textbf{EP} & \textbf{KTN (gain$\%$)} \\ \midrule\hline
\multirow{4}{*}{\textbf{P-A (L1)}} & \textbf{NDCG} & 0.399 & 0.452 & 0.405 & 0.292 & 0.262 & 0.261 & 0.26 & 0.178 & 0.425 & \textbf{0.623 (56)} \\
 & \textbf{std} & 0.010 & 0.012 & 0.032 & 0.009 & 0.021 & 0.014 & 0.021 & 0.000 & 0.006 & 0.004 \\
 & \textbf{MRR} & 0.297 & 0.361 & 0.314 & 0.179 & 0.129 & 0.111 & 0.138 & 0.041 & 0.363 & \textbf{0.629 (112)} \\
 & \textbf{std} & 0.024 & 0.006 & 0.041 & 0.011 & 0.032 & 0.031 & 0.033 & 0.000 & 0.005 & 0.004 \\ \hline
\multirow{4}{*}{\textbf{A-P (L1)}} & \textbf{NDCG} & 0.401 & 0.566 & 0.598 & 0.294 & 0.364 & 0.246 & 0.195 & 0.153 & 0.557 & \textbf{0.733 (83)} \\
 & \textbf{std} & 0.003 & 0.012 & 0.014 & 0.034 & 0.049 & 0.046 & 0.025 & 0.000 & 0.002 & 0.007 \\
 & \textbf{MRR} & 0.318 & 0.508 & 0.544 & 0.229 & 0.27 & 0.09 & 0.047 & 0.022 & 0.507 & \textbf{0.711 (123)} \\
 & \textbf{std} & 0.001 & 0.029 & 0.028 & 0.093 & 0.117 & 0.037 & 0.029 & 0.000 & 0.003 & 0.009 \\ \hline
\multirow{4}{*}{\textbf{A-V (L1)}} & \textbf{NDCG} & 0.459 & 0.457 & 0.47 & 0.382 & 0.346 & 0.359 & 0.403 & 0.207 & 0.461 & \textbf{0.671 (46)} \\
 & \textbf{std} & 0.030 & 0.033 & 0.036 & 0.015 & 0.029 & 0.109 & 0.024 & 0.000 & 0.002 & 0.004 \\
 & \textbf{MRR} & 0.364 & 0.413 & 0.458 & 0.341 & 0.205 & 0.253 & 0.327 & 0.011 & 0.389 & \textbf{0.698 (92)} \\
 & \textbf{std} & 0.079 & 0.08 & 0.093 & 0.05 & 0.098 & 0.143 & 0.044 & 0.000 & 0.004 & 0.003 \\ \hline
\multirow{4}{*}{\textbf{V-A (L1)}} & \textbf{NDCG} & 0.283 & 0.443 & 0.435 & 0.242 & 0.372 & 0.418 & 0.272 & 0.153 & 0.154 & \textbf{0.584 (107)} \\
 & \textbf{std} & 0.045 & 0.012 & 0.007 & 0.004 & 0.048 & 0.039 & 0.004 & 0.000 & 0.006 & 0.005 \\
 & \textbf{MRR} & 0.133 & 0.365 & 0.345 & 0.094 & 0.241 & 0.444 & 0.144 & 0.006 & 0.006 & \textbf{0.586 (340)} \\
 & \textbf{std} & 0.074 & 0.027 & 0.017 & 0.011 & 0.103 & 0.115 & 0.018 & \multicolumn{1}{l}{} & 0.007 & 0.010 \\ \hline
\multirow{4}{*}{\textbf{P-A (L2)}} & \textbf{NDCG} & 0.229 & 0.23 & o.o.m & 0.239 & o.o.m & o.o.m & 0.168 & o.o.m & 0.215 & \textbf{0.282 (23)} \\
 & \textbf{std} & 0.005 & 0.003 & - & 0.006 & - & - & 0.007 & - & 0.004 & 0.002 \\
 & \textbf{MRR} & 0.121 & 0.118 & o.o.m & 0.14 & o.o.m & o.o.m & 0.02 & o.o.m & 0.143 & \textbf{0.2248 (86)} \\
 & \textbf{std} & 0.019 & 0.004 & - & 0.01 & - & - & 0.006 & - & 0.003 & 0.003 \\ \hline
\multirow{4}{*}{\textbf{A-P (L2)}} & \textbf{NDCG} & 0.197 & 0.162 & o.o.m & 0.204 & 0.158 & 0.161 & 0.132 & o.o.m & 0.208 & \textbf{0.287 (46)} \\
 & \textbf{std} & 0.006 & 0.009 & - & 0.006 & 0.019 & 0.022 & 0.012 & - & 0.004 & 0.001 \\
 & \textbf{MRR} & 0.095 & 0.052 & o.o.m & 0.106 & 0.032 & 0.045 & 0.017 & o.o.m & 0.132 & \textbf{0.242 (155)} \\
 & \textbf{std} & 0.009 & 0.022 & - & 0.016 & 0.018 & 0.027 & 0.008 & - & 0.005 & 0.002 \\ \hline
\multirow{4}{*}{\textbf{A-V (L2)}} & \textbf{NDCG} & 0.347 & 0.329 & 0.295 & 0.325 & 0.288 & 0.273 & 0.289 & o.o.m & 0.297 & \textbf{0.402 (16)} \\
 & \textbf{std} & 0.003 & 0.034 & 0.014 & 0.013 & 0.011 & 0.058 & 0.011 & - & 0.002 & 0.003 \\
 & \textbf{MRR} & 0.310 & 0.296 & 0.198 & 0.223 & 0.128 & 0.097 & 0.11 & o.o.m & 0.227 & \textbf{0.399 (29)} \\
 & \textbf{std} & 0.004 & 0.109 & 0.047 & 0.065 & 0.003 & 0.096 & 0.034 & - & 0.001 & 0.015 \\ \hline
\multirow{4}{*}{\textbf{V-A (L2)}} & \textbf{NDCG} & 0.235 & 0.249 & 0.251 & 0.214 & 0.197 & 0.205 & 0.217 & o.o.m & 0.119 & \textbf{0.252 (7)} \\
 & \textbf{std} & 0.002 & 0.002 & 0.006 & 0.004 & 0.008 & 0.004 & 0.002 & - & 0.001 & 0.007 \\
 & \textbf{MRR} & 0.130 & 0.157 & 0.161 & 0.09 & 0.044 & 0.068 & 0.085 & o.o.m & 0.000 & \textbf{0.166 (28)} \\
 & \textbf{std} & 0.010 & 0.011 & 0.009 & 0.015 & 0.007 & 0.009 & 0.005 & - & 0.000 & 0.012 \\ \hline\bottomrule
\end{tabular}
\vspace{5mm}
\end{table*}

\subsection{More results for Zero-shot Transfer Learning in Section~\ref{sec:experiments:zero-shot}}
\label{appendix:result:da}

We show the zero-shot transfer learning results with error bars on OAG-computer science and Pubmed in Tables~\ref{tab:oag:cs2} and~\ref{tab:pubmed2}.
We also present the results with error bars on OAG-computer networks and OAG-machine learning in Tables~\ref{tab:oag:cn} and~\ref{tab:oag:ml}, respectively.
Across all tasks and graphs, our proposed method \method consistently outperforms all baselines.

\subsection{Analysis for Baselines in Section~\ref{sec:experiments:zero-shot}}
\label{appendix:analysis}

Among baselines, MMD-based models (DAN and JAN) outperform adversarial based methods (DANN, CDAN, and CDAN-E) and optimal transport-based method (WDGRL), unlike results reported in~\cite{long2017conditional, shen2018wasserstein}.
These reversed results are a consequence of HGNN's unique feature extractors for source and target domains.
When $f_s$ and $f_t$ denote feature extractors for source and target domains, respectively, DANN and CDAN define their adversarial losses as a cross entropy loss ($\mathbb{E}[\log f_s] - \mathbb{E}[\log f_t]$) where gradients of the subloss $\mathbb{E}[\log f_s]$ are passed only back to $f_s$, while gradients of the subloss $\mathbb{E}[\log f_t]$ are passed only back to $f_t$.
Importantly, source and target feature extractors do not share any gradient information, resulting in divergence.
This did not occur in their original test environments where source and target domains share a single feature extractor.
Similarly, WDGRL measures the first-order Wasserstein distance as an adversarial loss, which also brings the same effect as the cross-entropy loss we described above, leading to divergent gradients between source and target feature extractors.
On the other hand, DAN and JAN define a loss in terms of higher-order MMD between source and target features.
Then the gradients of the loss passed to each feature extractor contain both source and target feature information, resulting in a more stable gradient estimation.
This shows again the importance of considering different feature extractors in HGNNs.

JAN, CDAN, and CDAN-E often show out of memory issues in Tables~\ref{tab:oag:cs2},~\ref{tab:oag:cn}, and~\ref{tab:oag:ml}.
These baselines consider the classifier prediction whose dimension is equal to the number of classes in a given task.
That is why JAN, CDAN, and CDAN-E fail at the L2 field prediction tasks in OAG graphs where the number of classes is $17,729$.

LP performs worst among the baselines, showing the limitation of relying only on graph structures.
LP maintains a label vector with the length equal to the number of classes for each node, thus shows out-of-memory issues on tasks with large number of classes on large-size graphs (L2 tasks with $17,729$ labels on the OAG-CS graph).
EP performs moderately well similar to other DA methods, but lower than \method up to $60\%$ absolute points of MRR, showing the limitation of not using target node attributes.

\begin{table*}[]
    \caption{
    \small
	\textbf{PubMed}
	}
	\label{tab:pubmed2}
	\centering
    \tiny
\begin{tabular}{l|l|c|cc|ccc|c|cc|l} \toprule\hline
\textbf{Task} & \textbf{Metric} & \textbf{Base} & \textbf{DAN} & \textbf{JAN} & \textbf{DANN} & \textbf{CDAN} & \textbf{CDAN-E} & \textbf{WDGRL} & \textbf{LP} & \textbf{EP} & \textbf{KTN (gain$\%$)} \\ \midrule\hline
\multirow{4}{*}{\textbf{D-G}} & \textbf{NDCG} & 0.587 & 0.629 & 0.615 & 0.614 & 0.624 & 0.646 & 0.604 & 0.601 & 0.571 & \textbf{0.700 (19)} \\
 & \textbf{std} & 0.004 & 0.013 & 0.028 & 0.008 & 0.078 & 0.015 & 0.022 & 0.000 & 0.004 & 0.005 \\
 & \textbf{MRR} & 0.372 & 0.425 & 0.414 & 0.397 & 0.428 & 0.443 & 0.388 & 0.389 & 0.336 & \textbf{0.499 (34)} \\
 & \textbf{std} & 0.003 & 0.007 & 0.054 & 0.013 & 0.066 & 0.027 & 0.035 & 0.000 & 0.003 & 0.006 \\ \hline
\multirow{4}{*}{\textbf{G-D}} & \textbf{NDCG} & 0.596 & 0.599 & 0.577 & 0.599 & 0.581 & 0.606 & 0.578 & 0.576 & 0.580 & \textbf{0.662 (11)} \\
 & \textbf{std} & 0.007 & 0.020 & 0.032 & 0.011 & 0.054 & 0.019 & 0.019 & 0.000 & 0.011 & 0.003 \\
 & \textbf{MRR} & 0.354 & 0.362 & 0.332 & 0.356 & 0.337 & 0.362 & 0.340 & 0.351 & 0.353 & \textbf{0.445 (26)} \\
 & \textbf{std} & 0.005 & 0.015 & 0.019 & 0.008 & 0.023 & 0.031 & 0.015 & 0.000 & 0.008 & 0.002 \\ \hline\bottomrule
\end{tabular}
\end{table*}

\begin{table*}[]
    \caption{
	\textbf{Open Academic Graph on Computer Network field}
	}
	\label{tab:oag:cn}
	\centering
    \tiny
\begin{tabular}{l|l|c|cc|ccc|c|cc|l} \toprule\hline
\textbf{Task} & \textbf{Metric} & \textbf{Base} & \textbf{DAN} & \textbf{JAN} & \textbf{DANN} & \textbf{CDAN} & \textbf{CDAN-E} & \textbf{WDGRL} & \textbf{LP} & \textbf{EP} & \textbf{KTN (gain$\%$)} \\ \midrule\hline
\multirow{4}{*}{\textbf{P-A (L2)}} & \textbf{NDCG} & 0.331 & 0.344 & o.o.m & 0.335 & o.o.m & o.o.m & 0.287 & 0.221 & 0.270 & \textbf{0.382 (16)} \\
 & \textbf{std} & 0.004 & 0.005 & - & 0.004 & - & - & 0.012 & 0.000 & 0.003 & 0.004 \\
 & \textbf{MRR} & 0.250 & 0.277 & o.o.m & 0.280 & o.o.m & o.o.m & 0.199 & 0.130 & 0.270 & \textbf{0.360 (44)} \\
 & \textbf{std} & 0.024 & 0.012 & - & 0.007 & - & - & 0.004 & 0.000 & 0.003 & 0.010 \\ \hline
\multirow{4}{*}{\textbf{A-P (L2)}} & \textbf{NDCG} & 0.313 & 0.290 & o.o.m & 0.250 & 0.234 & 0.168 & 0.266 & 0.114 & 0.319 & \textbf{0.364 (17)} \\
 & \textbf{std} & 0.002 & 0.023 & - & 0.021 & 0.041 & 0.025 & 0.030 & 0.000 & 0.004 & 0.003 \\
 & \textbf{MRR} & 0.250 & 0.233 & o.o.m & 0.130 & 0.116 & 0.051 & 0.212 & 0.038 & 0.296 & \textbf{0.368 (47)} \\
 & \textbf{std} & 0.015 & 0.039 & - & 0.051 & 0.069 & 0.037 & 0.061 & 0.000 & 0.005 & 0.004 \\ \hline
\multirow{4}{*}{\textbf{A-V (L2)}} & \textbf{NDCG} & 0.539 & 0.521 & 0.519 & 0.510 & 0.467 & 0.362 & 0.471 & 0.232 & 0.443 & \textbf{0.567 (5)} \\
 & \textbf{std} & 0.012 & 0.031 & 0.008 & 0.022 & 0.008 & 0.045 & 0.024 & 0.000 & 0.002 & 0.008 \\
 & \textbf{MRR} & 0.584 & 0.528 & 0.461 & 0.510 & 0.293 & 0.294 & 0.365 & 0.000 & 0.406 & \textbf{0.628 (8)} \\
 & \textbf{std} & 0.042 & 0.015 & 0.011 & 0.054 & 0.013 & 0.088 & 0.019 & 0.000 & 0.004 & 0.016 \\ \hline
\multirow{4}{*}{\textbf{V-A (L2)}} & \textbf{NDCG} & 0.256 & 0.343 & 0.345 & 0.265 & 0.328 & 0.316 & 0.263 & 0.133 & 0.119 & \textbf{0.341 (33)} \\
 & \textbf{std} & 0.006 & 0.012 & 0.005 & 0.005 & 0.005 & 0.003 & 0.003 & 0.000 & 0.001 & 0.005 \\
 & \textbf{MRR} & 0.117 & 0.296 & 0.286 & 0.151 & 0.285 & 0.275 & 0.147 & 0.000 & 0.000 & \textbf{0.281 (141)} \\
 & \textbf{std} & 0.020 & 0.009 & 0.004 & 0.009 & 0.006 & 0.008 & 0.009 & 0.000 & 0.000 & 0.014 \\\hline\bottomrule
\end{tabular}
\vspace{5mm}
\end{table*}

\subsection{More results for Generality of \method in Section~\ref{sec:experiments:hgnn-types}}
\label{appendix:result:hgnn}

We show \method performance on $6$ different types of HGNN models across $4$ different zero-shot domain adaptation tasks on the OAG-computer science dataset in Table~\ref{tab:hgnn2}.
Descriptions of each HGNN model can be found in Appendix~\ref{appendix:hgnn_models}.
While \method consistently improves all HGNN models' performance on zero-labeled node types using labels rooted at other node types, the magnitude of improvements varies.
While HAN sees up to $4958\%$ (V-A (L1) task in Table~\ref{tab:hgnn2}), MAGNN is improved by up to $47\%$ (P-A(L1) task) or sees no improvement (A-V(L1) task).
This gap stems from how many parameters each HGNN model shares across node types.
HAN does not share any parameters during message-passing operations (every parameters are specialized to each meta-path), while MAGNN shares the transformation matrices across all node types at every layer.
By sharing more parameters with other node types, the gradients are more likely passed to target node type-specific parameters, leaving less room for improvement by \method.
However, \method is still necessary for any HGNN models.
MPNN who shares all parameters except the projection matrices that map different input attributes into the same embedding space at the beginning still sees improvements by up to $311\%$.
Again, these experimental results show the impact of having different feature extractors for each node type in HGNN models.

\begin{table*}[]
    \caption{
	\textbf{Open Academic Graph on Machine Learning field}
	}
	\label{tab:oag:ml}
	\centering
    \tiny
\begin{tabular}{l|l|c|cc|ccc|c|cc|l} \toprule\hline
\textbf{Task} & \textbf{Metric} & \textbf{Base} & \textbf{DAN} & \textbf{JAN} & \textbf{DANN} & \textbf{CDAN} & \textbf{CDAN-E} & \textbf{WDGRL} & \textbf{LP} & \textbf{EP} & \textbf{KTN (gain$\%$)} \\ \midrule\hline
\multirow{4}{*}{\textbf{P-A (L2)}} & \textbf{NDCG} & 0.268 & 0.290 & o.o.m & 0.291 & o.o.m & 0.249 & 0.232 & 0.272 & 0.215 & \textbf{0.318 (19)} \\
 & \textbf{std} & 0.002 & 0.009 & - & 0.004 & - & 0.005 & 0.004 & 0.000 & 0.002 & 0.004 \\
 & \textbf{MRR} & 0.134 & 0.220 & o.o.m & 0.222 & o.o.m & 0.095 & 0.098 & 0.195 & 0.143 & \textbf{0.269 (102)} \\
 & \textbf{std} & 0.006 & 0.020 & - & 0.026 & - & 0.003 & 0.037 & 0.000 & 0.003 & 0.006 \\ \hline
\multirow{4}{*}{\textbf{A-P (L2)}} & \textbf{NDCG} & 0.261 & 0.225 & o.o.m & 0.234 & 0.228 & 0.241 & 0.241 & 0.119 & 0.267 & \textbf{0.319 (22)} \\
 & \textbf{std} & 0.002 & 0.009 & - & 0.004 & 0.005 & 0.011 & 0.002 & 0.000 & 0.001 & 0.005 \\
 & \textbf{MRR} & 0.207 & 0.127 & o.o.m & 0.155 & 0.152 & 0.095 & 0.182 & 0.035 & 0.214 & \textbf{0.287 (39)} \\
 & \textbf{std} & 0.018 & 0.042 & - & 0.008 & 0.009 & 0.003 & 0.017 & 0.000 & 0.012 & 0.011 \\ \hline
\multirow{4}{*}{\textbf{A-V (L2)}} & \textbf{NDCG} & 0.465 & 0.493 & 0.463 & 0.477 & 0.408 & 0.422 & 0.393 & 0.224 & 0.424 & \textbf{0.538 (16)} \\
 & \textbf{std} & 0.006 & 0.004 & 0.003 & 0.003 & 0.006 & 0.013 & 0.005 & 0.000 & 0.005 & 0.004 \\
 & \textbf{MRR} & 0.469 & 0.542 & 0.537 & 0.519 & 0.412 & 0.240 & 0.213 & 0.001 & 0.391 & \textbf{0.632 (35)} \\
 & \textbf{std} & 0.039 & 0.008 & 0.005 & 0.003 & 0.015 & 0.008 & 0.009 & 0.000 & 0.021 & 0.006 \\ \hline
\multirow{4}{*}{\textbf{V-A (L2)}} & \textbf{NDCG} & 0.252 & 0.293 & 0.292 & 0.237 & 0.242 & 0.255 & 0.250 & 0.137 & 0.119 & \textbf{0.302 (20)} \\
 & \textbf{std} & 0.006 & 0.011 & 0.009 & 0.004 & 0.003 & 0.002 & 0.004 & 0.000 & 0.003 & 0.007 \\
 & \textbf{MRR} & 0.131 & 0.212 & 0.199 & 0.086 & 0.085 & 0.129 & 0.118 & 0.000 & 0.000 & \textbf{0.227 (73)} \\
 & \textbf{std} & 0.016 & 0.023 & 0.013 & 0.005 & 0.021 & 0.007 & 0.012 & 0.000 & 0.000 & 0.015 \\\hline\bottomrule
\end{tabular}
\end{table*}

\begin{table*}[t]
    \caption{
	\textbf{Meta-path length in \method:}
	increasing the meta-path longer than the minimum does not bring significant improvement to \method.
	Note that the minimum length of meta-paths in the A-V (L1) task is $2$.
	}
	\label{tab:meta-path}
	\centering
    \small
\begin{tabular}{c|llll}
    \toprule \hline
    \textbf{Task}                                                    & \multicolumn{2}{c}{\textbf{P-A (L1)}}             & \multicolumn{2}{c}{\textbf{A-V (L1)}} \\ \hline
    \textbf{\begin{tabular}[c]{@{}c@{}}Meta-path\\ length\end{tabular}} & \textbf{NDCG} & \multicolumn{1}{l|}{\textbf{MRR}} & \textbf{NDCG}      & \textbf{MRR}     \\ \hline \midrule
    \textbf{1}                                                       & 0.623         & \multicolumn{1}{l|}{0.621}        & 0.208              & 0.010            \\
    \textbf{2}                                                       & 0.627         & \multicolumn{1}{l|}{0.628}        & 0.673              & 0.696            \\
    \textbf{3}                                                       & 0.608         & \multicolumn{1}{l|}{0.611}        & 0.627              & 0.648            \\
    \textbf{4}                                                       & 0.61          & \multicolumn{1}{l|}{0.623}        & 0.653              & 0.671            \\
    \hline \bottomrule
\end{tabular}
\vspace{5mm}
\end{table*}

\begin{table*}[]
    \caption{
	\textbf{\method on different HGNN models}: 
	The \textit{Source} column shows accuracy on source node types. 
	\textit{Base} and \textit{\method} columns show accuracy on target node types without/with using \method, respectively.
	The \textit{Gain} column shows the relative gain of our method over using no transfer learning.
	}
	\label{tab:hgnn2}
	\centering
    \tiny
\begin{tabular}{l|l|ccc|r|ccc|r} \toprule\hline
\textbf{} &  & \multicolumn{4}{c}{\textbf{P-A (L1)}} & \multicolumn{4}{c}{\textbf{A-P (L1)}} \\
\textbf{HGNN type} & \textbf{Metric} & \textbf{Source} & \textbf{Base} & \textbf{KTN} & \textbf{Gain$\%$} & \textbf{Source} & \textbf{Base} & \textbf{KTN} & \textbf{Gain$\%$} \\ \midrule\hline
\multirow{4}{*}{\textbf{R-GCN}} & \textbf{NDCG} & 0.765 & 0.337 & 0.577 & \textbf{71.12} & 0.648 & 0.388 & 0.647 & \textbf{66.82} \\
 & \textbf{std} & 0.004 & 0.005 & 0.002 & \multicolumn{1}{l|}{\textbf{}} & 0.006 & 0.007 & 0.004 & \multicolumn{1}{l}{\textbf{}} \\
 & \textbf{MRR} & 0.757 & 0.236 & 0.587 & \textbf{148.73} & 0.623 & 0.270 & 0.611 & \textbf{126.18} \\
 & \textbf{std} & 0.002 & 0.003 & 0.001 & \multicolumn{1}{l|}{\textbf{}} & 0.005 & 0.008 & 0.004 & \multicolumn{1}{l}{\textbf{}} \\ \hline
\multirow{4}{*}{\textbf{HAN}} & \textbf{NDCG} & 0.476 & 0.179 & 0.520 & \textbf{190.56} & 0.515 & 0.182 & 0.512 & \textbf{181.33} \\
 & \textbf{std} & 0.004 & 0.006 & 0.003 & \multicolumn{1}{l|}{\textbf{}} & 0.004 & 0.009 & 0.011 & \multicolumn{1}{l}{\textbf{}} \\
 & \textbf{MRR} & 0.416 & 0.047 & 0.497 & \textbf{960.55} & 0.509 & 0.055 & 0.527 & \textbf{850.90} \\
 & \textbf{std} & 0.001 & 0.002 & 0.002 & \multicolumn{1}{l|}{\textbf{}} & 0.005 & 0.004 & 0.005 & \multicolumn{1}{l}{\textbf{}} \\ \hline
\multirow{4}{*}{\textbf{HGT}} & \textbf{NDCG} & 0.757 & 0.294 & 0.574 & \textbf{95.07} & 0.670 & 0.283 & 0.581 & \textbf{104.83} \\
 & \textbf{std} & 0.002 & 0.003 & 0.004 & \multicolumn{1}{l|}{} & 0.001 & 0.003 & 0.009 & \multicolumn{1}{l}{} \\
 & \textbf{MRR} & 0.749 & 0.178 & 0.563 & \textbf{216.17} & 0.670 & 0.149 & 0.565 & \textbf{279.52} \\
 & \textbf{std} & 0.005 & 0.007 & 0.001 & \multicolumn{1}{l|}{\textbf{}} & 0.002 & 0.007 & 0.006 & \multicolumn{1}{l}{\textbf{}} \\ \hline
\multirow{4}{*}{\textbf{MAGNN}} & \textbf{NDCG} & 0.657 & 0.463 & 0.574 & \textbf{24.01} & 0.676 & 0.557 & 0.622 & \textbf{11.68} \\
 & \textbf{std} & 0.003 & 0.001 & 0.003 & \multicolumn{1}{l|}{\textbf{}} & 0.001 & 0.001 & 0.003 & \multicolumn{1}{l}{\textbf{}} \\
 & \textbf{MRR} & 0.631 & 0.378 & 0.556 & \textbf{47.33} & 0.680 & 0.509 & 0.592 & \textbf{16.14} \\
 & \textbf{std} & 0.003 & 0.002 & 0.004 & \multicolumn{1}{l|}{\textbf{}} & 0.001 & 0.002 & 0.005 & \multicolumn{1}{l}{\textbf{}} \\ \hline
\multirow{4}{*}{\textbf{MPNN}} & \textbf{NDCG} & 0.602 & 0.443 & 0.590 & \textbf{33.11} & 0.646 & 0.307 & 0.621 & \textbf{101.92} \\
 & \textbf{std} & 0.002 & 0.002 & 0.001 & \multicolumn{1}{l|}{\textbf{}} & 0.005 & 0.013 & 0.004 & \multicolumn{1}{l}{\textbf{}} \\
 & \textbf{MRR} & 0.572 & 0.319 & 0.575 & \textbf{80.10} & 0.660 & 0.145 & 0.595 & \textbf{311.42} \\
 & \textbf{std} & 0.001 & 0.003 & 0.005 & \multicolumn{1}{l|}{} & 0.002 & 0.024 & 0.003 & \multicolumn{1}{l}{} \\ \hline
\multirow{4}{*}{\textbf{H-MPNN}} & \textbf{NDCG} & 0.789 & 0.399 & 0.623 & \textbf{56.14} & 0.671 & 0.401 & 0.733 & \textbf{82.88} \\
 & \textbf{std} & 0.001 & 0.005 & 0.001 & \multicolumn{1}{l|}{\textbf{}} & 0.003 & 0.005 & 0.009 & \multicolumn{1}{l}{\textbf{}} \\
 & \textbf{MRR} & 0.777 & 0.297 & 0.629 & \textbf{111.86} & 0.661 & 0.318 & 0.711 & \textbf{123.30} \\
 & \textbf{std} & 0.003 & 0.001 & 0.002 & \multicolumn{1}{l|}{} & 0.007 & 0.004 & 0.008 & \multicolumn{1}{l}{} \\ \hline\midrule \toprule
\textbf{} & \multicolumn{1}{c}{\textbf{}} & \multicolumn{4}{c}{\textbf{V-A (L1)}} & \multicolumn{4}{c}{\textbf{A-V (L1)}} \\ 
\textbf{HGNN type} & \textbf{Metric} & \textbf{Source} & \textbf{Base} & \textbf{KTN} & \textbf{Gain$\%$} & \textbf{Source} & \textbf{Base} & \textbf{KTN} & \textbf{Gain$\%$} \\ \midrule\hline
\multirow{4}{*}{\textbf{R-GCN}} & \textbf{NDCG} & 0.664 & 0.426 & 0.530 & \textbf{24.36} & 0.660 & 0.599 & 0.744 & \textbf{24.26} \\
 & \textbf{std} & 0.003 & 0.006 & 0.002 & \multicolumn{1}{l|}{\textbf{}} & 0.001 & 0.008 & 0.004 & \multicolumn{1}{l}{\textbf{}} \\
 & \textbf{MRR} & 0.683 & 0.325 & 0.514 & \textbf{58.39} & 0.656 & 0.524 & 0.785 & \textbf{49.87} \\
 & \textbf{std} & 0.003 & 0.008 & 0.004 & \multicolumn{1}{l|}{\textbf{}} & 0.011 & 0.009 & 0.005 & \multicolumn{1}{l}{\textbf{}} \\ \hline
\multirow{4}{*}{\textbf{HAN}} & \textbf{NDCG} & 0.618 & 0.153 & 0.510 & \textbf{232.35} & 0.515 & 0.546 & 0.689 & \textbf{26.21} \\
 & \textbf{std} & 0.005 & 0.007 & 0.003 & \multicolumn{1}{l|}{\textbf{}} & 0.008 & 0.003 & 0.005 & \multicolumn{1}{l}{\textbf{}} \\
 & \textbf{MRR} & 0.634 & 0.010 & 0.516 & \textbf{4958.82} & 0.508 & 0.511 & 0.758 & \textbf{48.28} \\
 & \textbf{std} & 0.002 & 0.005 & 0.002 & \multicolumn{1}{l|}{\textbf{}} & 0.001 & 0.008 & 0.007 & \multicolumn{1}{l}{\textbf{}} \\ \hline
\multirow{4}{*}{\textbf{HGT}} & \textbf{NDCG} & 0.615 & 0.234 & 0.536 & \textbf{128.95} & 0.694 & 0.367 & 0.735 & \textbf{100.22} \\
 & \textbf{std} & 0.002 & 0.005 & 0.002 & \multicolumn{1}{l|}{\textbf{}} & 0.006 & 0.007 & 0.009 & \multicolumn{1}{l}{\textbf{}} \\
 & \textbf{MRR} & 0.638 & 0.095 & 0.537 & \textbf{464.88} & 0.699 & 0.267 & 0.772 & \textbf{189.21} \\
 & \textbf{std} & 0.006 & 0.002 & 0.005 &  & 0.002 & 0.005 & 0.012 &  \\ \hline
\multirow{4}{*}{\textbf{MAGNN}} & \textbf{NDCG} & 0.536 & 0.513 & 0.513 & \textbf{0.00} & 0.684 & 0.676 & 0.692 & \textbf{2.37} \\
 & \textbf{std} & 0.005 & 0.001 & 0.001 & \textbf{} & 0.001 & 0.002 & 0.001 & \textbf{} \\
 & \textbf{MRR} & 0.586 & 0.522 & 0.522 & \textbf{0.00} & 0.686 & 0.751 & 0.752 & \textbf{0.13} \\
 & \textbf{std} & 0.004 & 0.001 & 0.002 & \textbf{} & 0.002 & 0.001 & 0.004 & \textbf{} \\ \hline
\multirow{4}{*}{\textbf{MPNN}} & \textbf{NDCG} & 0.578 & 0.380 & 0.532 & \textbf{40.03} & 0.639 & 0.578 & 0.794 & \textbf{37.19} \\
 & \textbf{std} & 0.008 & 0.008 & 0.004 & \multicolumn{1}{l|}{\textbf{}} & 0.007 & 0.007 & 0.005 & \multicolumn{1}{l}{\textbf{}} \\
 & \textbf{MRR} & 0.603 & 0.253 & 0.505 & \textbf{100.12} & 0.652 & 0.584 & 0.847 & \textbf{44.96} \\
 & \textbf{std} & 0.001 & 0.003 & 0.007 & \multicolumn{1}{l|}{\textbf{}} & 0.006 & 0.001 & 0.006 & \multicolumn{1}{l}{\textbf{}} \\ \hline
\multirow{4}{*}{\textbf{H-MPNN}} & \textbf{NDCG} & 0.670 & 0.283 & 0.584 & \textbf{106.50} & 0.676 & 0.459 & 0.671 & \textbf{46.22} \\
 & \textbf{std} & 0.002 & 0.002 & 0.006 & \multicolumn{1}{l|}{\textbf{}} & 0.005 & 0.004 & 0.003 & \multicolumn{1}{l}{\textbf{}} \\
 & \textbf{MRR} & 0.689 & 0.133 & 0.586 & \textbf{339.76} & 0.677 & 0.364 & 0.698 & \textbf{91.92} \\
 & \textbf{std} & 0.003 & 0.003 & 0.005 & \multicolumn{1}{l|}{\textbf{}} & 0.01 & 0.005 & 0.002 & \multicolumn{1}{l}{\textbf{}} \\ \hline\bottomrule
\end{tabular}
\vspace{5mm}
\end{table*}

\subsection{Effect of trade-off coefficient $\lambda$}
\label{sec:experiments:lambda}

We examine the effect of $\lambda$ on transfer learning performance.
In Table~\ref{tab:lambda}, as $\lambda$ decreases, target accuracy decreases as expected.
Source accuracy also sees small drops since $\mathcal{L}_{\text{KTN}}$ functions as a regularizer; by removing the regularization effect, source accuracy decreases.
When $\lambda$ becomes large, both source and target accuracy drop significantly.
Source accuracy drops since the effect of $\mathcal{L}_{\text{KTN}}$ becomes larger than the classification loss $\mathcal{L}_{\text{CL}}$.
Even the effect of transfer learning become larger by having larger $\lambda$, since the source accuracy which will be transferred to the target domain is low, the target accuracy is also low. 
Thus we set $\lambda$ to $1$ throughout the experiments.

\begin{table*}[]
    \caption{
	\textbf{Effect of $\lambda$}
	}
	\label{tab:lambda}
	\centering
    \tiny
\begin{tabular}{c|cccc|cccc}
\toprule\hline
\textbf{}          & \multicolumn{4}{c|}{\textbf{P-A (L1)}}                                                     & \multicolumn{4}{c}{\textbf{A-V (L1)}}                                                      \\
\textbf{Metric}    & \multicolumn{2}{c|}{\textbf{NDCG}}                     & \multicolumn{2}{c|}{\textbf{MRR}} & \multicolumn{2}{c|}{\textbf{NDCG}}                     & \multicolumn{2}{c}{\textbf{MRR}}  \\ \hline
$\lambda$             & \textbf{Source} & \multicolumn{1}{l|}{\textbf{Target}} & \textbf{Source} & \textbf{Target} & \textbf{Source} & \multicolumn{1}{l|}{\textbf{Target}} & \textbf{Source} & \textbf{Target} \\ \hline
$10^{-5}$      & 0.780           & \multicolumn{1}{l|}{0.587}           & 0.772           & 0.595           & 0.689           & \multicolumn{1}{l|}{0.626}           & 0.690           & 0.642           \\
$10^{-3}$     & 0.788           & \multicolumn{1}{l|}{0.58}            & 0.779           & 0.576           & 0.687           & \multicolumn{1}{l|}{0.654}           & 0.689           & 0.677           \\
$10^0$      & 0.792           & \multicolumn{1}{l|}{0.621}           & 0.788           & 0.633           & 0.689           & \multicolumn{1}{l|}{0.670}            & 0.692           & 0.696           \\
$10^2$    & 0.750            & \multicolumn{1}{l|}{0.617}           & 0.757           & 0.623           & 0.654           & \multicolumn{1}{l|}{0.644}           & 0.659           & 0.668           \\
$10^4$ & 0.143           & \multicolumn{1}{l|}{0.177}           & 0.007           & 0.031           & 0.411           & \multicolumn{1}{l|}{0.432}           & 0.373           & 0.421          \\ \hline \bottomrule
\end{tabular}
\normalsize
\end{table*}

\subsection{Synthetic Heterogeneous Graph Generator}
\label{appendix:graph-generator}

Our synthetic heterogeneous graph generator is based on attributed Stochastic Block Models (SBM)~\cite{tsitsulin2020graph,tsitsulin2021synthetic}, using blocks (clusters) as the node classes.
In the attributed SBM, graphs exhibit \emph{within-type} cluster homophily at the \emph{edge-level} (nodes most-frequently connect to other nodes in the same cluster), and at the \emph{feature-level} (nodes are closest in feature space to other nodes in the same cluster). 
To produce heterogeneous graphs, we additionally introduce \emph{between-type} cluster homophily, which allows us to model real-world heterogeneous graphs in which knowledge can be shared across node types. 

The first step in generating a heterogeneous SBM is to decide how many clusters will partition each node type. Assume \emph{within-type} cluster counts $k_1, \ldots, k_T$. We allow for \emph{between-type} cluster homophily with a $K_T:=\min_t\{k_t\}$-partition of clusters such that each cluster group has at least one corresponding cluster from other node types.

Secondly, edge-level homophily is controlled by signal-to-noise ratios $\sigma_e = p/q$ where nodes \emph{within-cluster} are connected with probability $p$ and nodes \emph{between-cluster} are connected with probability $q$. Additionally, edges \emph{within one cluster group across different types} (see previous paragraph) is controlled together with edges \emph{between different cluster groups across different types} using some $\sigma_e$.
In Section~\ref{sec:experiments:sensitivity}, we describe the manipulation of multiple $\sigma_e$ parameters \emph{within-and-between} types.

Finally, node attributes are generated by a multivariate Normal mixture model, using the cluster partition as the mixture groups.
Thus feature-level homophily is controlled by increasing the variance of the cluster centers $\sigma_f$, while keeping the within-cluster variance fixed. 
Cross-type feature homophily is controlled by setting a center of cluster centers \emph{within-type} with linear combinations of centers (of cluster centers) of other types.
Note that features of different types are allowed to have different dimensions, as we generate different mixture-model cluster centers for each cluster \emph{within each type}. 

\subsubsection{Toy Heterogeneous Graph in Section~\ref{sec:motivation:experiments}}
\label{appendix:graph-generator:toy}

Using the synthetic graph procedure described above, we used the following hyperparameters to simulate the toy heterogeneous graph shown in Figure~\ref{fig:toy_exp}.
We generate the graph with $2$ node types and $4$ edge types as described in Figure~\ref{fig:toy:hg}, then we divide each node type into $4$ classes of $400$ nodes.
To generate an easy-to-transfer scenario, signal-to-noise ratio $\sigma_f$ between means of feature distributions are all set to $10$.
The ratio $\sigma_e$ of the number of intra-class edges to the number of inter-class edges is set to $10$ among the same node types and across different node types.
The dimension of features is set to $24$ for both node types.

\subsubsection{Sensitivity test in Section~\ref{sec:experiments:sensitivity}}
\label{appendix:graph-generator:sensitivity}

Figure~\ref{fig:schema2} shows the structures of graphs we used in Section~\ref{sec:experiments:sensitivity}.
The dimension of features are set to $24$ for both node types for the "easy" scenario, and $32$ and $48$ for types $s$ and $t$, respectively, for the "hard" scenario. Additionally, for the "hard" scenario, we divide the $t$ nodes into $8$ clusters instead of $4$.
The other hyperparameters $\sigma_e$ and $\sigma_f$ are described in Section~\ref{sec:experiments:sensitivity}.
For each unique value of $\sigma_{(\cdot)}$ across the six ($\sigma_{(\cdot)}, r$) pairs, we generate $5$ heterogeneous graphs.

\begin{table*}[t!]
    \caption{
	\textbf{Statistics of Open Academic Graph}
	}
	\label{tab:oag:statistics}
	\centering
\small
\begin{tabular}{l|l|l|l|l|l|l}
    \toprule\hline
    \textbf{Domain}           & \textbf{\#papers} & \textbf{\#authors} & \textbf{\#fields} & \textbf{\#venues} & \textbf{\#institues} &                \\ \hline\midrule
    \textbf{Computer Science} & 544,244           & 510,189            & 45,717            & 6,934             & 9,097                &                \\
    \textbf{Computer Network} & 75,015            & 82,724             & 12,014            & 2,115             & 4,193                &                \\
    \textbf{Machine Learning} & 90,012            & 109,423            & 19,028            & 3,226             & 5,455                &                \\ \hline
    \textbf{Domain}           & \textbf{\#P-A}    & \textbf{\#P-F}     & \textbf{\#P-V}    & \textbf{\#A-I}    & \textbf{\#P-P}       & \textbf{\#F-F} \\ \hline\midrule
    \textbf{Computer Science} & 1,091,560         & 3,709,711          & 544,245           & 612,873           & 11,592,709           & 525,053        \\
    \textbf{Computer Network} & 155,147           & 562,144            & 75,016            & 111,180           & 1,154,347            & 110,869        \\
    \textbf{Machine Learning} & 166,119           & 585,339            & 90,013            & 156,440           & 1,209,443            & 163,837        \\ \hline\bottomrule
\end{tabular}
\normalsize
\end{table*}

\begin{table*}[t!]
    \caption{
	\textbf{Statistics of PubMed Graph}
	}
	\label{tab:pubmed:statistics}
	\centering
\small
\begin{tabular}{p{1.5cm}|p{1.5cm}|p{1.5cm}|p{1.5cm}|p{1.5cm}}
\toprule\hline
\textbf{\#gene} & \textbf{\#disease} & \textbf{\#chemicals} & \textbf{\#species} &                \\ \hline\midrule
13,561           & 20,163             & 26,522                & 2,863               &                \\ \hline
\textbf{\#G-G}  & \textbf{\#G-D}     & \textbf{\#D-D}       & \textbf{\#C-G}     & \textbf{\#C-D} \\ \hline\midrule
32,211           & 25,963              & 68,219                & 31,278              & 51,324          \\ \hline
\textbf{\#C-C}  & \textbf{\#C-S}     & \textbf{\#S-G}       & \textbf{\#S-D}     & \textbf{\#S-S} \\ \hline\midrule
124,375          & 6,298               & 3,156                 & 5,246               & 1,597          \\ \hline\bottomrule
\end{tabular}
\vspace{5mm}
\end{table*}

\subsection{Real-world Dataset}
\label{appendix:dataset}

\paragraph{Open Academic Graph (OAG)}~\cite{sinha2015overview, tang2008arnetminer, zhang2019oag} is the largest publicly available heterogeneous graph.
It is composed of five types of nodes: papers, authors, institutions, venues, fields and their corresponding relationships.
Papers and authors have text-based attributes, while institutions, venues, and fields have text- and graph structure-based attributes.
To test the generalization of the proposed model, we construct three field-specific subgraphs from OAG: the Computer
Science (OAG-CS), Computer Networks (OAG-CN), and Machine Learning (OAG-ML) academic graphs. 

Papers, authors, and venues are labeled with research fields in two hierarchical levels, L1 and L2.
OAG-CS has both L1 and L2 labels, while OAG-CN and OAG-ML have only L2 labels (their L1 labels are all "computer science").
Transfer learning is performed on the L1 and L2 field prediction tasks between papers, authors, and venues for each of the aforementioned subgraphs.
Note that paper-author (P-A) and paper-venue (P-V) are directly connected, while author-venue (A-V) are indirectly connected via papers.

\begin{figure}[t!]
 	\centering
 	\subfigure[Synthetic graph]
 	{
 	\label{fig:schema2}
 	\includegraphics[width=0.24\linewidth]{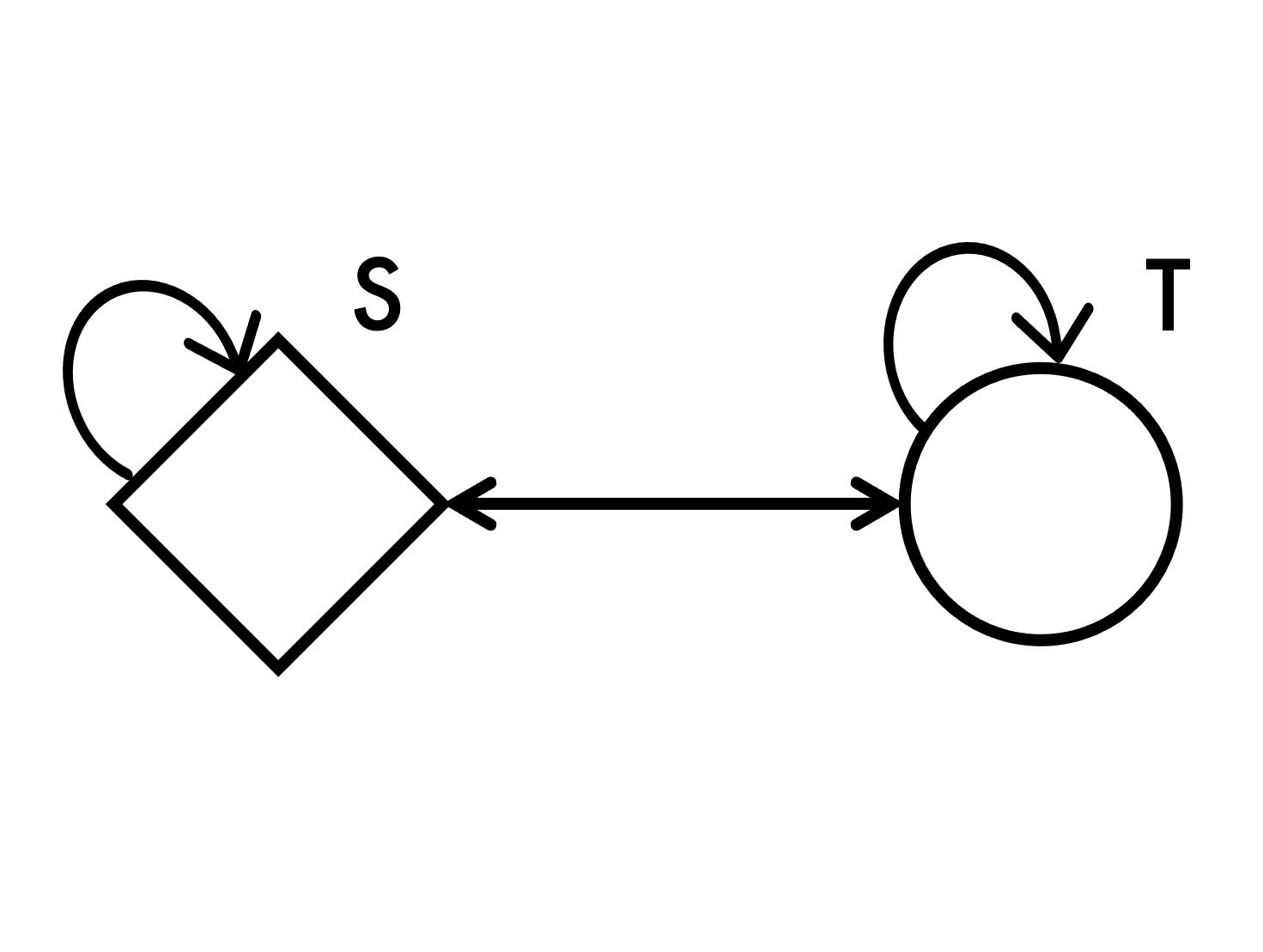}
 	}
 	\subfigure[OAG]
 	{
 	\label{fig:schema1:oag}
 	\includegraphics[width=0.38\linewidth]{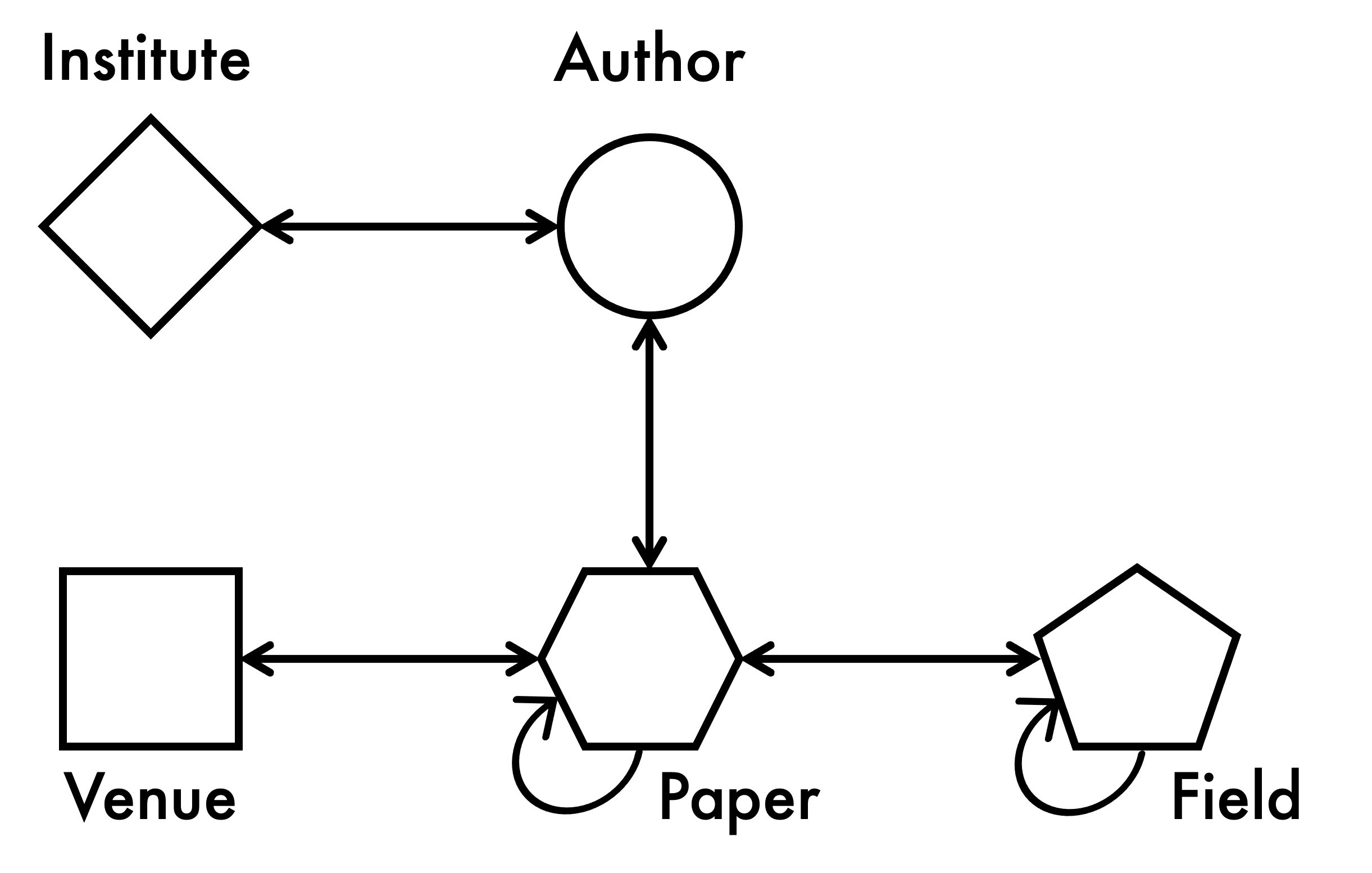}
 	}
 	\subfigure[PubMed]
 	{
 	\label{fig:schema1:pubmed}
 	\includegraphics[width=.27\linewidth]{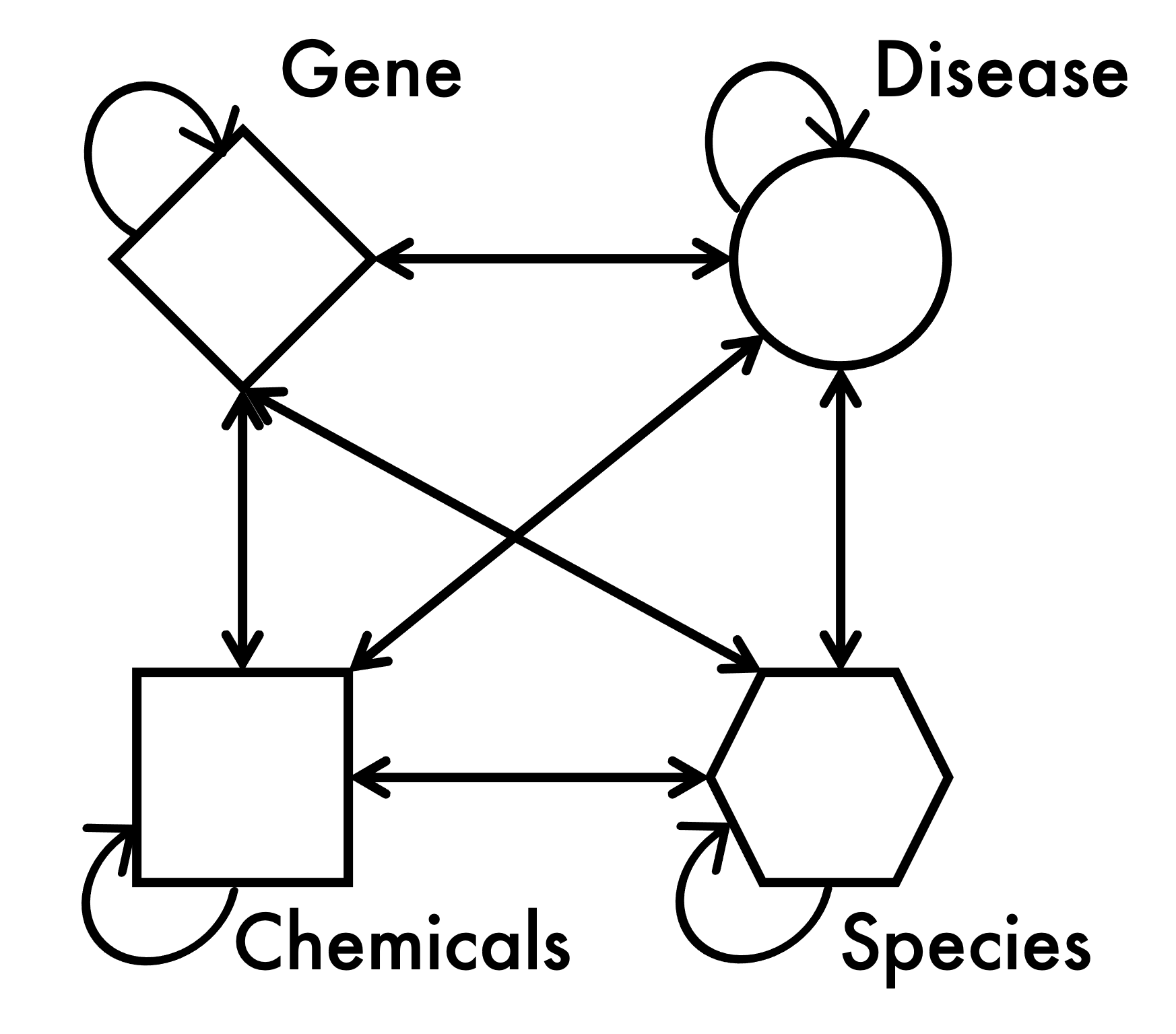}
 	}
 	\caption
 	{
 	    Schema of synthetic and real-world heterogeneous graphs
 	}
 	\label{fig:schema1}
 	\vspace{5mm}
 \end{figure}

The number of classes in the L1 task is $275$, while the number of classes in the L2 task is $17,729$.
The graph statistics are listed in Table~\ref{tab:oag:statistics}, in which P–A, P–F, P–V, A–I, P–P, and F-F denote the edges between paper and author, paper and field, paper and venue, author and institute, the citation links between two papers, and the hierarchical links between two fields.
The graph structure is described in Figure~\ref{fig:schema1:oag}.

For paper nodes, features are generated from each paper's title using a pre-trained XLNet~\cite{wolf2020transformers}.
For author nodes, features are averaged over features of papers they published.
Feature dimension of paper and author nodes is $769$.
For venue, institution, and field node types, features of dimension $400$ are generated from their heterogeneous graph structures using metapath2vec~\cite{dong2017metapath2vec}.

\paragraph{PubMed}~\cite{yang2020heterogeneous} is a novel biomedical network constructed through text mining and manual processing on biomedical literature.
PubMed is composed of genes, diseases, chemicals, and species.
Each gene or disease is labeled with a set of diseases (e.g., cardiovascular disease) they belong to or cause.
Transfer learning is performed on a disease prediction task between genes and disease node types. 

The number of classes in the disease prediction task is $8$.
The graph statistics are listed in Table~\ref{tab:pubmed:statistics}, in which G, D, C, and S denote genes, diseases, chemicals, and species node types.
The graph structure is described in Figure~\ref{fig:schema1:pubmed}.

For gene and chemical nodes, features of dimension $200$ are generated from related PubMed papers using word2vec~\cite{mikolov2013distributed}.
For diseases and species nodes, features of dimension $50$ are generated based on their graph structures using TransE~\cite{bordes2013translating}.

\subsection{Baselines}
\label{appendix:baseline}

Zero-shot domain adaptation can be categorized into three groups --- MMD-based methods, adversarial methods, and optimal-transport-based methods.
MMD-based methods~\cite{long2015learning, long2017deep, sun2016return} minimize the maximum mean discrepancy (MMD)~\cite{gretton2012kernel} between the mean embeddings of two distributions in reproducing kernel Hilbert space.
DAN~\cite{long2015learning} enhances the feature transferability by minimizing multi-kernel MMD in several task-specific layers.
JAN~\cite{long2017deep} aligns the joint distributions of multiple domain-specific layers based on a joint maximum mean discrepancy (JMMD) criterion.

Adversarial methods~\cite{ganin2016domain, long2017conditional} are motivated by theory in~\cite{ben2010theory, ben2007analysis} suggesting that a good cross-domain representation contains no discriminative information about the origin of the input.
They learn domain invariant features by a min-max game between the domain classifier and the feature extractor.
DANN~\cite{ganin2016domain} learns domain invariant features by a min-max game between the domain classifier and the feature extractor.
CDAN~\cite{long2017conditional} exploits discriminative information conveyed in the classifier predictions to assist adversarial adaptation.
CDAN-E~\cite{long2017conditional} extends CDAN to condition the domain discriminator on the uncertainty of classifier predictions, prioritizing the discriminator on easy-to-transfer examples.

Optimal transport-based methods~\cite{shen2018wasserstein} estimate the empirical Wasserstein distance~\cite{redko2017theoretical} between two domains and minimizes the distance in an adversarial manner.
Optimal transport-based method are based on a theoretical analysis~\cite{redko2017theoretical} that Wasserstein distance can guarantee generalization for domain adaptation.
WDGRL~\cite{shen2018wasserstein} estimates the empirical Wasserstein distance between two domains and minimizes the distance in an adversarial manner.

\subsection{HGNN models}
\label{appendix:hgnn_models}

We briefly describe $6$ heterogeneous graph neural networks (HGNN) models we used in the experiments.
MPNN (message passing neural networks)~\cite{gilmer2017neural} is originally designed for homogeneous graphs.
We extend MPNN to process heterogeneous graphs by adding projection matrices that project input attributes of different node types into the same feature space before running the original MPNN.  
R-GCN~\cite{schlichtkrull2018modeling} extends MPNN by specializing message matrices in each edge type, while HMPNN specializes all transformation and message matrices in each node/edge type in MPNN.
HGT~\cite{hu2020heterogeneous} extends HMPNN by adding attention modules.
The attention modules have node-type-specific key/query projection matrices and edge-type-specific key-query similarity matrices, following the transformer architecture.

HAN~\cite{wang2019heterogeneous} is a meta-path-based model who specializes parameters in each meta-path.
HAN exploits meta-path-specific attention modules to aggregate features of neighboring nodes connected by each meta-path.
Then HAN aggregates embeddings of different meta-paths with semantic-level attention modules.
MAGNN~\cite{fu2020magnn} is another meta-path-based HGNN model.
MAGNN aggregates features of all nearby nodes sitting on each meta-path using intra-meta-path attention modules.
Then MAGNN aggregates features of different meta-paths using inter-meta-path attention modules.

All HGNN models we describe above have layer-wise parameters.
As all HGNN models have parameters specialized in either node/edge/meta-path types, they all have distinct feature extractors for each node types, thus, they will suffer from the under-trained target node phenomena we showed in Section~\ref{sec:motivation:experiments}.
Also, because the core intuition in \method — namely that embeddings of any node types at the last layer are computed using the same set of the previous layer’s intermediate embeddings (see Section~\ref{sec:motivation:theoretical_analysis}) — holds across all HGNN models, \method can be applied to any HGNN models and show greatly increased target-type accuracy.

\subsection{Experimental Settings}
\label{appendix:experiment-setting}

All experiments were conducted on the same p2.xlarge Amazon EC2 instance.
Here, we describe the structure of HGNNs used in each heterogeneous graph.

\paragraph{Open Academic Graph:}
We use a $4$-layered HGNN with transformation and message parameters of dimension $128$ for \method and other baselines.
Learning rate is set to $10^{-4}$.

\paragraph{PubMed:}
We use a single-layered HGNN with transformation and message parameters of dimension $10$ for \method and other baselines.
Learning rate is set to $5 \times 10^{-5}$.

\paragraph{Synthetic Heterogeneous Graphs:}
We use a $2$-layered HGNN with transformation and message parameters of dimension $128$ for \method and other baselines.
Learning rate is set to $10^{-4}$.

We implement LP, EP and \method using Pytorch.
For the domain adaptation baselines (DAN, JAN, DANN, CDAN, CDAN-E, and WDGRL), we use a public domain adaptation library ADA~\footnote{\url{https://github.com/criteo-research/pytorch-ada}}.
We match the numbers of layers and dimensions of hidden embeddings across all HGNN models.
We implement MPNN and HMPNN using Pytorch.
For other HGNN models (R-GCN, HAN, HGT, and MAGNN), we use an open-source toolkit for Heterogeneous Graph Neural Network (OpenHGNN)~\footnote{\url{https://github.com/BUPT-GAMMA/OpenHGNN}}.
Our code is publicly available~\footnote{\url{https://github.com/minjiyoon/KTN}}.

\end{document}